\documentclass[10pt,twocolumn,letterpaper]{article}

\usepackage{iccv}
\usepackage{times}
\usepackage{epsfig}
\usepackage{graphicx}
\usepackage{amsmath}
\usepackage{amssymb}

\usepackage[sort&compress,numbers]{natbib}
\usepackage{booktabs}
\usepackage{multirow}
\usepackage{multicol}
\usepackage{subcaption}
\usepackage{tabularx}
\usepackage[font=small,labelfont=small]{caption}
\usepackage{soul}
\usepackage[accsupp]{axessibility}

\usepackage{amsmath,amsfonts,bm}

\def\eqref#1{equation~\ref{#1}}

\def\1{\bm{1}}

\def\rvg{{\mathbf{g}}}
\def\rvh{{\mathbf{h}}}

\def\rvs{{\mathbf{s}}}

\def\rvw{{\mathbf{w}}}
\def\rvx{{\mathbf{x}}}
\def\rvy{{\mathbf{y}}}

\def\vs{{\bm{s}}}

\DeclareMathAlphabet{\mathsfit}{\encodingdefault}{\sfdefault}{m}{sl}
\SetMathAlphabet{\mathsfit}{bold}{\encodingdefault}{\sfdefault}{bx}{n}

\def\gG{{\mathcal{G}}}

\newcommand{\KL}{D_{\mathrm{KL}}}

\newcommand{\normlone}{L^1}

\newcommand\blfootnote[1]{%
  \begingroup
  \renewcommand\thefootnote{}\footnote{#1}%
  \addtocounter{footnote}{-1}%
  \endgroup
}

\usepackage[pagebackref=true,breaklinks=true,letterpaper=true,colorlinks,bookmarks=false]{hyperref}

\iccvfinalcopy %

\def\iccvPaperID{10917} %

\def\eg{\emph{e.g.}}

\def\ie{\emph{i.e.}}

\def\etc{\emph{etc.}}
\def\vs{\emph{vs.}}

\ificcvfinal\fi

\begin{document}

\title{{LayoutTransformer: Layout Generation and Completion with Self-attention}}

\author{
$\textbf{Kamal Gupta}^{1\star},\quad \textbf{Justin Lazarow}^{2},\quad \textbf{Alessandro Achille}^{3}$,\\[0.2em]
$\textbf{Larry Davis}^{1,3},\quad \textbf{Vijay Mahadevan}^{3\star},\quad \textbf{Abhinav Shrivastava}^{1}$
\\[0.5em]
$^{1}\text{University of Maryland, College Park}$\quad
$^{2}\text{University of California, San Diego}$\quad
$^{3}\text{Amazon AWS}$
}

\maketitle
\ificcvfinal\thispagestyle{empty}\fi

\blfootnote{$^{\star}$ Corresponding authors.\newline\hspace*{15pt}Work started during an internship at Amazon.}

\begin{abstract}
We address the problem of scene layout generation for diverse domains such as images, mobile applications, documents, and 3D objects. Most complex scenes, natural or human-designed, can be expressed as a meaningful arrangement of simpler compositional graphical primitives. Generating a new layout or extending an existing layout requires understanding the relationships between these primitives. To do this, we propose \textbf{LayoutTransformer}, a novel framework that leverages self-attention to learn contextual relationships between layout elements and generate novel layouts in a given domain. Our framework allows us to generate a new layout either from an empty set or from an initial seed set of primitives, and can easily scale to support an arbitrary of primitives per layout. Furthermore, our analyses show that the model is able to automatically capture the semantic properties of the primitives. We propose simple improvements in both representation of layout primitives, as well as training methods to demonstrate competitive performance in very diverse data domains such as object bounding boxes in natural images (COCO bounding box), documents (PubLayNet), mobile applications (RICO dataset) as well as 3D shapes (PartNet). Code and other materials will be made available at \url{https://kampta.github.io/layout}.
\end{abstract}

\section{Introduction}
\label{sec:introduction}

In the real world, there exists a strong relationship between different objects that are found in the same environment~\citep{torralba2001statistical,shrivastava2016contextual}. For example, a dining table usually has chairs around it, a surfboard is found near the sea, horses do not ride cars \etc\  \cite{biederman1981semantics} provided strong evidence in cognitive neuroscience that perceiving and understanding a scene involves two related processes: \emph{perception} and \emph{comprehension}. Perception deals with processing the visual signal or the appearance of a scene. Comprehension deals with understanding the \emph{schema} of a scene, where this schema (or layout) can be characterized by contextual relationships (\eg, support, occlusion, and relative likelihood, position, and size) between objects. For generative models that synthesize scenes, this evidence underpins the importance of two factors that contribute to the \emph{realism} or plausibility of a generated scene: layout, \ie, arrangement of different objects, and their appearance (in terms of pixels). Generating a realistic scene necessitates both the factors to be plausible.

\begin{figure}[!t]
\centering
  \includegraphics[width=\linewidth]{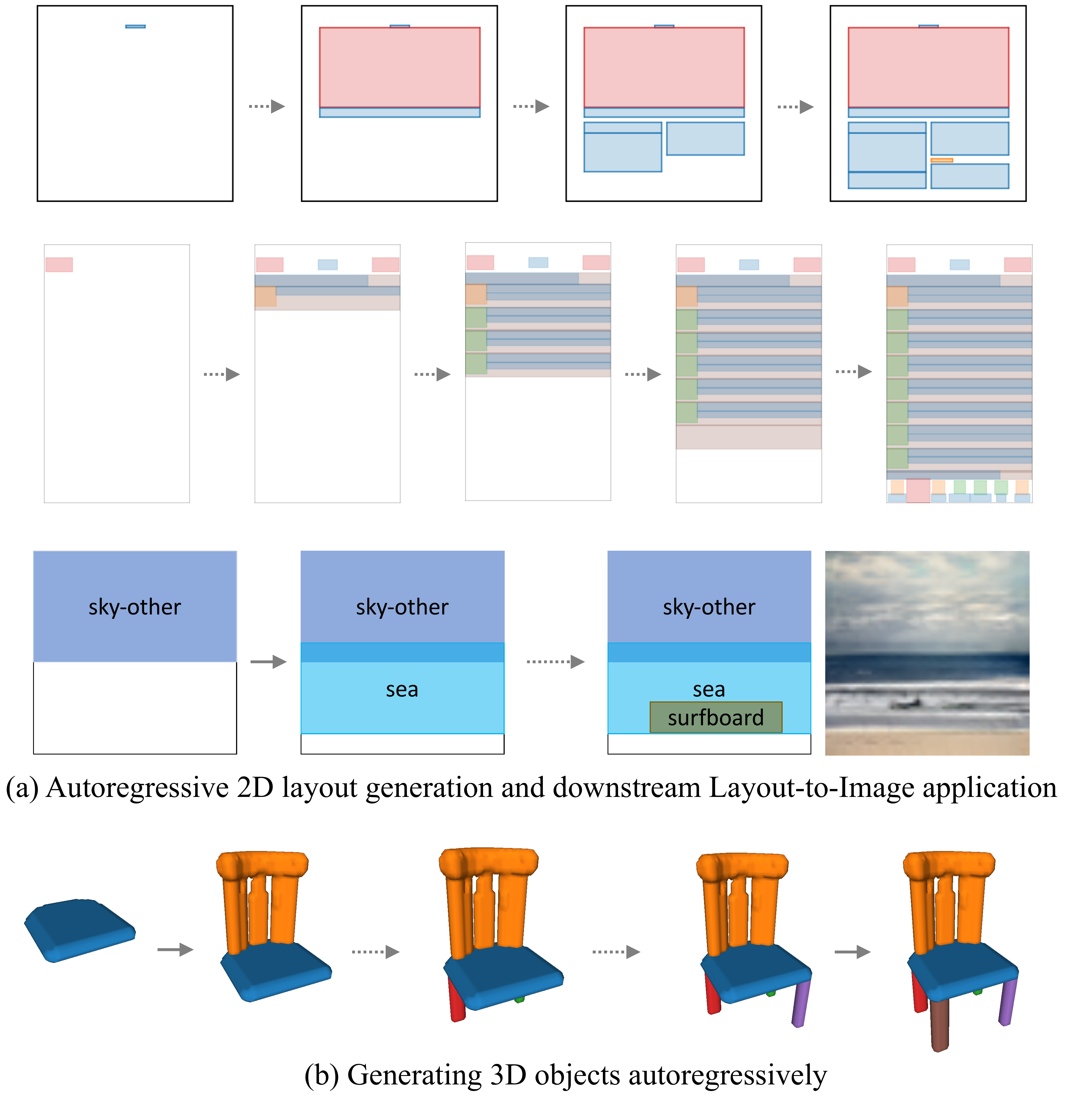}
\vspace{-0.2in}
\caption{Our framework can synthesize layouts in diverse natural as well as human designed data domains such as documents, mobile app wireframes, natural scenes or 3D objects in a sequential manner.}
\vspace{-0.2in}
\label{fig:teaser}
\end{figure}

The advancements in the generative models for image synthesis have primarily targeted plausibility of the appearance signal by generating incredibly realistic images often with a single entity such as faces~\citep{karras2019style,karras2017progressive}, or animals~\citep{brock2018large,zhang2018self}. 
In the case of large and complex scenes, with strong non-local relationships between different elements, most methods require proxy representations for layouts to be provided as inputs (\eg, scene graph, segmentation mask, sentence). We argue that to plausibly generate large scenes without such proxies, it is necessary to understand and generate the layout of a scene, in terms of contextual relationships between various objects present in the scene.

Learning to generate layouts is useful for several stand-alone applications that require generating layouts or templates with/without user interaction. For instance, in the UI design of mobile apps and websites, an automated model for generating plausible layouts can significantly decrease the manual effort and cost of building such apps and websites. Finally, a model to create layouts can potentially help generate synthetic data for various tasks tasks~\citep{DBLP:journals/corr/YangYAKKG17,DBLP:journals/corr/abs-1710-03474,DBLP:journals/corr/ChangMSPM15,nsd,vda}. Fig.~\ref{fig:teaser} shows some of the layouts autoregressively generated by our approach in various domains such as documents, mobile apps, natural scenes, and 3D shapes.

Formally, a scene layout can be represented as an unordered set of graphical primitives. The primitive itself can be discrete or continuous depending on the data domain. For example, in the case of layout of documents, primitives can be bounding boxes from discrete classes such as `text', `image', or `caption', and in case of 3D objects, primitives can be 3D occupancy grids of parts of the object such as `arm', `leg', or `back' in case of chairs. Additionally, in order to make the primitives compositional, we represent each primitive by a location vector with respect to the origin, and a scale vector that defines the bounding box enclosing the primitive. Again, based on the domain, these location and scale vectors can be 2D or 3D. A generative model for layouts should be able to look at all existing primitives and propose the placement and attributes of a new one. We propose a novel framework LayoutTransformer that first maps the different parameters of the primitive independently to a fixed-length continuous latent vector, followed by a masked Transformer decoder to look at representations of existing primitives in layout and predict the next primitive (one parameter at a time). Our generative framework can start from an empty set, or a set of primitives, and can iteratively generate a new primitive one parameter at a time. Moreover, by predicting either to stop or to generate the next primitive, our approach can generate variable length layouts. Our \textbf{main contributions} can be summarized as follows:

\begin{itemize}
    \item  We propose LayoutTransformer a simple yet powerful auto-regressive model that can synthesize new layouts, complete partial layouts, and compute likelihood of existing layouts. Self-attention approach allows us to visualize what existing elements are important for generating the next category in the sequence.
    \item We model different attributes of layout elements separately - doing so allows the attention module to more easily focus on the attributes that matter. This is important especially in datasets with inherent symmetries such as documents or apps and in contrast with existing approaches which concatenate or fuse different attributes of layout primitives.
    \item  We present an exciting finding -- encouraging a model to understand layouts results in feature representations that capture the semantic relationships between objects automatically (without explicitly using semantic embeddings, like word2vec~\citep{mikolov2013efficient}). This demonstrates the utility of the task of layout generation as a proxy-task for learning semantic representations, 
    \item LayoutTransformer shows good performance with essentially the same architecture and hyperpameters across very diverse domains. We show the adaptability of our model on four layout datasets: MNIST Layout~\cite{li2019layoutgan}, Rico Mobile App Wireframes~\citep{Deka:2017:Rico}, PubLayNet Documents~\cite{yepespublaynet}, and COCO Bounding Boxes~\citep{lin2014microsoft}. To the best of our knowledge, MMA is the first framework to perform competitively with the state-of-the-art approaches in 4 diverse data domains.
\end{itemize}

\section{Related Work}
\label{sec:related}
\vspace{-0.05in}
\noindent\textbf{Generative models.}
Deep generative models based on CNNs such as variational auto-encoders (VAEs) \citep{kingma2013autoencoding}, and generative adversarial networks (GANs)~\citep{goodfellow2014generative} have recently shown a great promise in terms of faithfully learning a given data distribution and sampling from it. There has also been research on generating data sequentially~\citep{van2016pixel,chen2020generative} even when the data has no natural order~\citep{vinyals2015order}. Many of these approaches often rely on low-level information~\citep{gupta2020patchvae} such as pixels while generating images~\citep{brock2018large,karras2019style}, videos~\citep{vondrick2016generating}, or 3D objects~\citep{wu2016learning,yang2019pointflow,park2019deepsdf,gupta2020improved} and not on semantic and geometric structure in the data.

\noindent\textbf{Scene generation.}
Generating 2D or 3D scenes conditioned on sentence~\citep{li2019object,zhang2017stackgan,reed2016generative}, a scene graph~\citep{johnson2018image,li2019seq,ashual2019specifying}, a layout~\citep{dong2017semantic,DBLP:journals/corr/abs-1901-00686,pix2pix2016,wang2018pix2pixHD} or an existing image~\citep{DBLP:journals/corr/abs-1812-02350} has drawn a great interest in vision community. Given the input, some works generate a fixed layout and diverse scenes~\citep{zhaobo2019layout2im}, while other works generate diverse layouts and scenes~\citep{johnson2018image,li2019object}. These methods involve pipelines often trained and evaluated end-to-end, and surprisingly little work has been done to evaluate the layout generation component itself. Layout generation serves as a complementary task to these works and can be combined with these methods. 
In this work, we evaluate the layout modeling capabilities of two of the recent works~\citep{johnson2018image,li2019object} that have layout generation as an intermediate step. We also demonstrate the results of our model with Layout2Im~\citep{zhaobo2019layout2im} for image generation.

\begin{figure*}[!ht]
\centering
    \includegraphics[trim={0 6cm 11cm 0}, clip, width=0.8\linewidth]{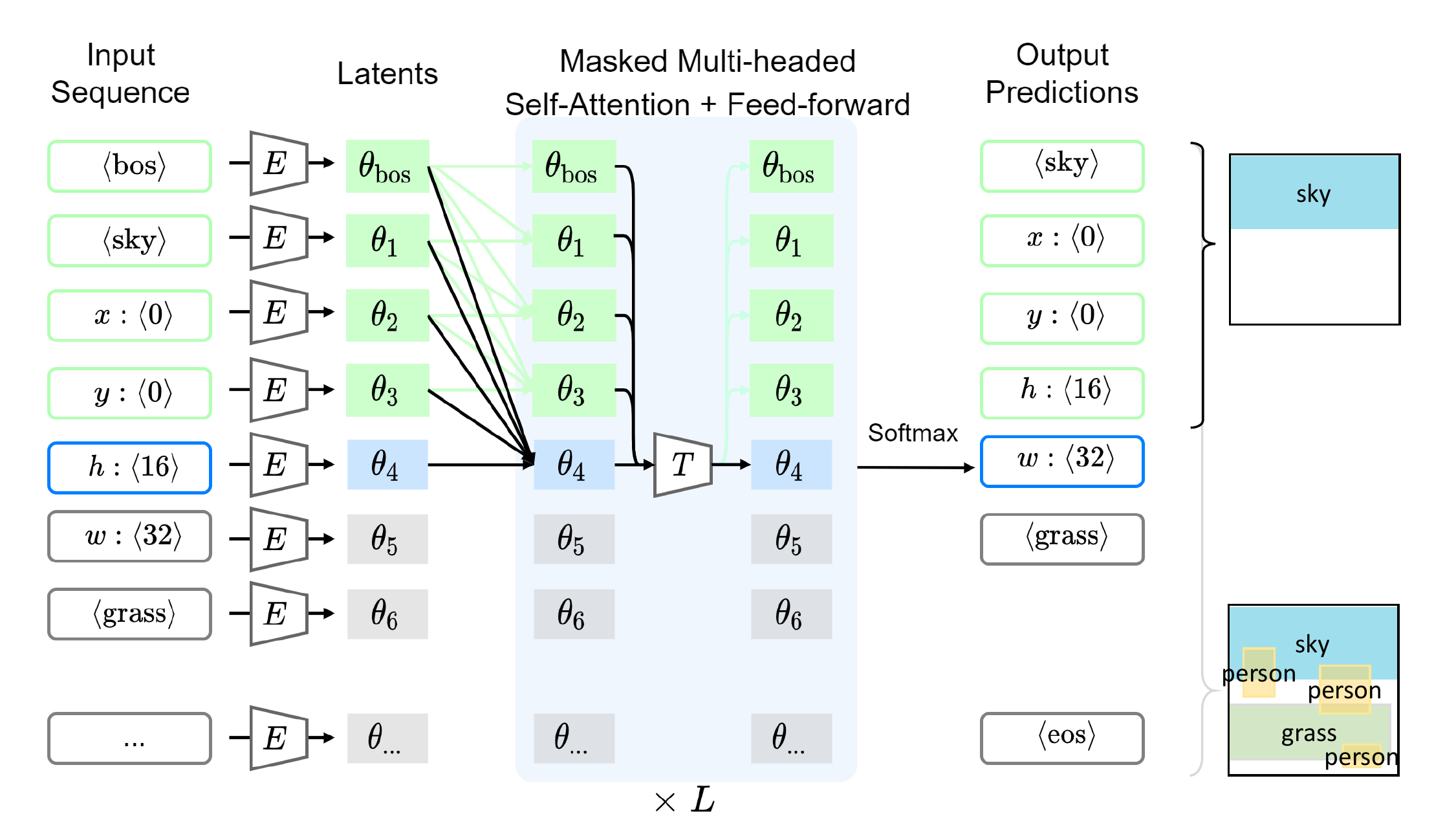}
\vspace{-0.2in}
  \caption{
  The architecture depicted for a toy example. LayoutTransformer takes layout elements as input and predicts the next layout elements as output. During training, we use teacher forcing, \textit{i.e.}, use the ground-truth layout tokens as input to a multi-head decoder block. The first layer of this block is a masked self-attention layer, which allows the model to see only the previous elements in order to predict the current element. We pad each layout with a special $\langle\text{bos}\rangle$ token in the beginning and $\langle\text{eos}\rangle$ token in the end.}
\vspace{-0.12in}
\label{fig:architecture}
\end{figure*}

\noindent\textbf{Layout generation.}
The automatic generation of layouts is an important problem in graphic design. Many of the recent data-driven approaches use data specific constraints in order to model the layouts. For example, ~\cite{wang2018deep,wang2019planit,li2019grains,ritchie2019fast} generates top-down view indoor rooms layouts but make several assumptions regarding the presence of walls, roof \etc, and cannot be easily extended to other datasets. In this paper, we focus on approaches that have fewer domain-specific constraints. LayoutGAN~\citep{li2019layoutgan} uses a GAN framework to generate semantic and geometric properties of a fixed number of scene elements. LayoutVAE~\citep{jyothi2019layoutvae} starts with a label set, \ie, categories of all the elements present in the layout, and then generates a feasible layout of the scene. \cite{zheng2019content} attempt to generate document layouts given the images, keywords, and category of the document. \cite{patil2019read} proposes a method to construct hierarchies of document layouts using a recursive variational autoencoder and sample new hierarchies to generate new document layouts. \cite{gcncnneccv2020} develops an auto-encoding framework for layouts using Graph Networks.
3D-PRNN~\citep{zou20173d}, PQ-Net~\citep{wu2020pq} and ComplementMe~\cite{sung2017complementme}, generates 3D shapes via sequential part assembly. While 3D-PRNN generates only bounding boxes, PQ-Net and ComplementMe can synthesize complete 3D shapes starting with a partial or no input shape.

\smallskip
Our approach offers several advantages over current layout generation approaches without sacrificing their benefits. By factorizing primitives into structural parameters and compositional geometric parameters, we can generate high-resolution primitives using distributed representations and consequently, complete scenes. The autoregressive nature of the model allows us to generate layouts of arbitrary lengths as well as start with partial layouts. Further, modeling the position and size of primitives as discrete values (as discussed in $\S$\ref{sec:setup}) helps us realize better performance on datasets, such as documents and app wireframes, where bounding boxes of layouts are typically axis-aligned. We evaluate our method both quantitatively and qualitatively with state-of-the-art methods specific to each dataset and show competitive results in very diverse domains.

\section{Our Approach}
\label{sec:approach}

In this section, we introduce our attention network in the context of the layout generation problem. We first discuss our representation of layouts for primitives belonging to different domains. Next, we discuss the LayoutTransformer framework and show how we can leverage Transformers~\citep{vaswani2017attention} to model the probability distribution of layouts. MMA allows us to learn non-local semantic relationships between layout primitives and also gives us the flexibility to work with variable length layouts.

\subsection{Layout Representation}
\label{sec:setup}

Given a dataset of layouts, a single layout instance can be defined as a graph $\gG$ with $n$ nodes, where each node $i \in \{1,\dots,n\}$ is a graphical primitive. We assume that the graph is fully-connected, and let the attention network learn the relationship between nodes. The nodes can have structural or semantic information associated with them.
For each node, we project the information associated with it to a $d$-dimensional space represented by feature vector $\rvs_i$. Note that the information itself can be discrete (\eg, part category), continuous (\eg, color), or multidimensional vectors (\eg, signed distance function of the part) on some manifold. Specifically, in our ShapeNet experiments, we use an MLP to project part embedding to $d$-dimensional space, while in the 2D layout experiments, we use a learned $d$-dimensional category embedding which is equivalent to using an MLP with zero bias to project one-hot encoded category vectors to the latent space.

Each primitive also carries geometric information $\rvg_i$ which we factorize into a position vector and a scale vector. For the layouts in $\mathbb{R}^2$ such as images or documents, $\rvg_i = [x_i, y_i, h_i, w_i]$, where $(x, y)$ are the coordinates of the centroid of primitive and $(h, w)$ are the height and width of the bounding box containing the primitive, normalized with respect to the dimensions of the entire layout.

\noindent\textbf{Representing geometry with discrete variables.} We apply an 8-bit uniform quantization on each of the geometric fields and model them using Categorical distribution. Discretizing continuous signals is a practice adopted in previous works for image generation such as PixelCNN++ \citep{salimans1701pixelcnn}, however, to the best of our knowledge, it has been unexplored in the layout modeling task. We observe that even though discretizing coordinates introduces approximation errors, it allows us to express arbitrary distributions which we find particularly important for layouts with strong symmetries such as documents and app wireframes. 
We project each geometric field of the primitive independently to the same $d$-dimension, such that $i^\text{th}$ primitive in $\mathbb{R}^2$ can be represented as $(\rvs_i, \rvx_i, \rvy_i, \rvh_i, \rvw_i)$. We concatenate all the elements in a flattened sequence of their parameters. We also append embeddings of two additional parameters $\rvs_{\langle \text{bos} \rangle}$ and $\rvs_{\langle \text{eos} \rangle}$ to denote start \& end of sequence. Layout in $\mathbb{R}^2$ can now be represented by a sequence of $5n+2$ latent vectors.
\[
\gG = (\rvs_{\langle \text{bos} \rangle}; \rvs_1; \rvx_1; \rvy_1; \rvh_1; \rvw_1; \dots; \rvs_n; \rvx_n; \rvy_n; \rvh_n; \rvw_n; \rvs_{\langle \text{eos} \rangle})
\]

For brevity, we use $\boldsymbol{\theta}_j$, $j \in \{1,\dots,5n+2\} $ to represent any element in the above sequence. We can now pose the problem of modeling this joint distribution as product over series of conditional distributions using chain rule:\vspace{-0.1in}
\begin{align}
p(\boldsymbol{\theta}_{1:5n+2}) = \prod_{j=1}^{5n+2} p(\boldsymbol{\theta}_j|\boldsymbol{\theta}_{1:{j-1}})
\end{align}

\begin{figure*}[!ht]
\centering
  \includegraphics[width=\linewidth]{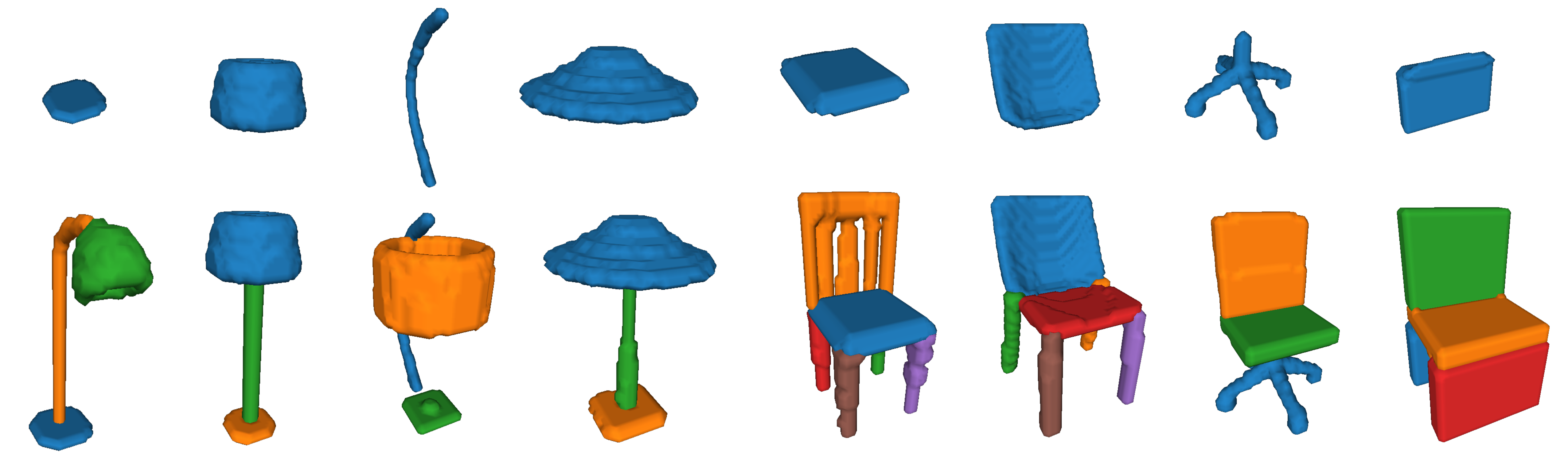}
  \vspace{-.25in}
  \caption{\textbf{Generated 3D objects.} Top row shows input primitives to the model. Bottom row shows the layout obtained by our approach.}
\label{fig:shape_generated}
\vspace{-.15in}
\end{figure*}

\subsection{Model architecture and training}
Our overall architecture is shown in Fig.~\ref{fig:architecture}.
Given an initial set of $K$ visible primitives (where $K$ can be $0$ when generating  from scratch), our attention based model takes as input, a random permutation of the visible nodes, $\boldsymbol{\pi} = (\pi_1,\dots,\pi_K)$, and consequently a sequence of $d$-dimensional vectors $(\boldsymbol{\theta}_1,\dots,\boldsymbol{\theta}_{5K})$. We find this to be an important step since by factorizing primitive representation into geometry and structure fields, our attention module can explicitly assign weights to individual coordinate dimensions. The attention module is similar to Transformer Decoder~\citep{vaswani2017attention} \& consists of $L$ attention layers, each comprising of (a) a masked multi-head attention layer ($\rvh^\text{attn}$), and (b) fully connected feed forward layer ($\rvh^\text{fc}$). Each sublayer also adds residual connections~\citep{he2016deep} and LayerNorm~\citep{ba2016layer}.
\begin{align}
    \hat{\boldsymbol{\theta}}_j &= \text{LayerNorm}(\boldsymbol{\theta}_j^{l-1} + \rvh^\text{attn}(\boldsymbol{\theta}_1^{l-1},\dots, \boldsymbol{\theta}_{j-1}^{l-1})) \\
    \boldsymbol{\theta}_j^l &= \text{LayerNorm}(\hat{\boldsymbol{\theta}}_j + \rvh^\text{fc}(\hat{\boldsymbol{\theta}}_j))
\end{align}
where $l$ denotes the layer index. Masking is performed such that $\boldsymbol{\theta}$ only attends to all the input latent vectors as well as previous predicted latent vectors. The output at the last layer corresponds to next parameter. At training and validation time, we use teacher forcing, \ie, instead of using output of previous step, we use groundtruth sequences to train our model efficiently.

\textbf{Loss.}
We use a softmax layer to get probabilities if the next parameter is discrete. 
Instead of using a standard cross-entropy loss, we minimize KL-Divergence between softmax predictions and output one-hot distribution with Label Smoothing~\citep{szegedy2016rethinking}, which prevents the model from becoming overconfident. If the next parameter is continuous, we use an $\normlone$ loss.
\begin{align*}
    \mathcal{L} &= \mathbb{E}_{\boldsymbol{\theta} \sim \text{Disc.}} [\ \KL ( \text{SoftMax}(\boldsymbol{\theta}^L) \ \Vert\  p(\boldsymbol{\theta}^{\prime}) )\ ] \\
    &+ \lambda 
    \mathbb{E}_{\boldsymbol{\theta} \sim \text{Cont.}}[\ || \boldsymbol{\theta} - \boldsymbol{\theta}^{\prime} ||_1\ ]
\end{align*}

\noindent\textbf{3D Primitive Auto-encoding.} PartNet dataset~\citep{yu2019partnet} consists of 3D objects decomposed into simpler meaningful primitives, such as chairs are composed of back, arms, 4 legs, and so on. We pose the problem of 3D shape generation as generating a layout of such primitives. We use~\cite{chen2019learning}'s approach to first encode voxel-based represent of primitive to $d$-dimensional latent space using 3D CNN. An MLP based implicit parameter decoder projects the latent vector to the surface occupancy grid of the primitive. 

\noindent\textbf{Order of primitives.} One of the limitations of an autoregressive modeling approach is that sequence of primitives is an important consideration, in order to train the generative model, even if the layout doesn't have a natural defined order~\cite{vinyals2015order}. To generate a layout from any partial layout, we use a random permutation of primitives as input to the model. For the output, we always generate the sequences in raster order of centroid of primitives, \ie, we order the primitives in ascending order of their $(x, y, z)$ coordinates. In our experiments, we observed that the ordering of elements is important for model training. Note that similar limitations are faced by contemporary works in layout generation ~\citep{jyothi2019layoutvae, li2019object, hong2018inferring, wang2018deep}, image generation~\citep{salimans1701pixelcnn,gregor2015draw} and 3D shape generation~\citep{wu2020pq,zou20173d}. Generating a distribution over an order-invariant set of an arbitrary number of primitives is an exciting problem and we will explore it in future research.

\noindent\textbf{Other details.} In our base model, we use $d=512$, $L=6$, and $n_\text{head}=8$ (number of multi-attention heads). Label smoothing uses an $\epsilon=0.1$, and $\lambda = 1$. We use Adam optimizer~\citep{kingma2014adam} with $\beta_1=0.9, \beta_2=0.99$ and learning rate $10^{-4}$ ($10^{-5}$ for PartNet). We use early stopping based on validation loss. In the ablation studies provided in appendix, we show that our model is quite robust to these choices, as well as other hyperparameters (layout resolution, ordering of elements, ordering of fields).
To sample a new layout, we can start off with just a start of sequence embedding or an initial set of primitives. Several decoding strategies are possible to recursively generate primitives from the initial set. In samples generated for this work, unless otherwise specified, we have used nucleus sampling~\citep{holtzman2019curious}, with top-$p = 0.9$ which has been shown to perform better as compared to greedy sampling and beam search~\citep{steinbiss1994improvements}.

\section{Experiments}
\label{sec:experiments}

In this section, we discuss the qualitative and quantitative performance of our model on different datasets. Evaluation of generative models is hard, and most quantitative measures fail in providing a good measure of novelty and realism of data sampled from a generative model. We will use dataset-specific quantitative metrics used by various baseline approaches and discuss their limitations wherever applicable. We will provide the code and pretrained models to reproduce the experiments.

\subsection{3D Shape synthesis (on PartNet dataset)}
PartNet is a large-scale dataset of common 3D shapes that are segmented into semantically meaningful parts. We use two of the largest categories of PartNet - Chairs and Lamp. We voxelize the shapes into $64^3$ and train an autoencoder to learn part embeddings similar to the procedure followed by PQ-Net~\citep{wu2020pq}. Overall, we had 6305 chairs and 1188 lamps in our datasets. We use the official train, validation, \& test split from PartNet. Although it is fairly trivial to extend our method to train for the class-conditional generation of shapes, in order to compare with baselines fairly, we train separate models for each of the categories.

\noindent\textbf{Generated Samples.}
Fig.~\ref{fig:shape_generated} shows examples of shape completion from the PartNet dataset. Given a random primitive, we use our model to iteratively predict the latent shape encoding of the next part, as well its position and scale in 3D. We then use the part decoder to sample points on the surface of the object. For visualization, we use the marching cubes algorithm to generate a mesh and render the mesh using a fixed camera viewpoint.

\begin{table*}[!h]
    
    \centering
    \caption{Evaluation of generated shapes in Chair category. The best numbers are in bold, second-best are underlined}
    \vspace{-0.1in}
    \renewcommand{\tabcolsep}{8pt}
    \renewcommand{\arraystretch}{1.1}
    \resizebox{\linewidth}{!}{
    \begin{tabular}{@{}lccccccc@{}}
        \toprule
        Method & JSD$\downarrow$ & MMD(CD)$\downarrow$ & MMD(EMD)$\downarrow$ & Cov(CD)$\uparrow$ & Cov(EMD)$\uparrow$ & 1-NNA(CD)$\downarrow$ & 1-NNA(EMD)$\downarrow$ \\
        \midrule
        PointFlow~\citep{yang2019pointflow}     & 1.74          & 2.42          &  \underline{7.87}          & 46.83          & 46.98          & \underline{60.88}         &  \underline{59.89} \\
        StructureNet~\citep{mo2019structurenet} & 4.77          & 0.97          &  15.24         & 29.67          & 31.7           & 75.32         &  74.22 \\
        IM-Net~\citep{chen2019learning}         & 0.84          & \textbf{0.74} &  12.28         & 52.35          & 54.12          & 68.52         &  67.12 \\
        PQ-Net~\citep{wu2020pq}                 & \underline{0.83} & 0.83          &  14.16         & \underline{54.91}          & \textbf{60.72} & 71.31         &  67.8 \\
        Ours                                   & \textbf{0.81}           & \underline{0.79}          &  \textbf{7.38} & \textbf{55.25} & \underline{55.44}          & \textbf{60.67}&  \textbf{59.11} \\
        \bottomrule
    \end{tabular}
    }
    \label{table:3dgen}
    \vspace{-0.2in}
\end{table*}

\noindent\textbf{Quantitative Evaluation.}
The output of our model is point clouds sampled on the surface of the 3D shapes. We use Chamfer Distance (CD) and Earth Mover's Distance (EMD) to compare two point clouds. Following prior work, we use 4 different metrics to compare the distribution of shapes generated from the model and shapes in the test dataset: (i) Jensen Shannon Divergence (JSD) computes the KL divergence between marginal distribution of point clouds in generated set and test set, (ii) Coverage (Cov) - compares the distance between each point in generated set to its nearest neighbor in test set, (iii) Minimum Matching Distance (MMD) - computes the average distance of each point in test set to its nearest neighbor in generated set, and (iv) 1-nearest neighbor accuracy (1-NNA) uses a 1-NN classifier see if the nearest neighbor of a generated sample is coming from generated set or test set. Our model performs competitively with existing approaches to generate point clouds. Table~\ref{table:3dgen} shows the generative performance of our model in the `Chair' category, with respect to recent proposed approaches. Our model's performance is either the best or second-best in all the metrics we evaluated in this work.

\begin{figure}[t]
    \centering
     \includegraphics[width=\linewidth]{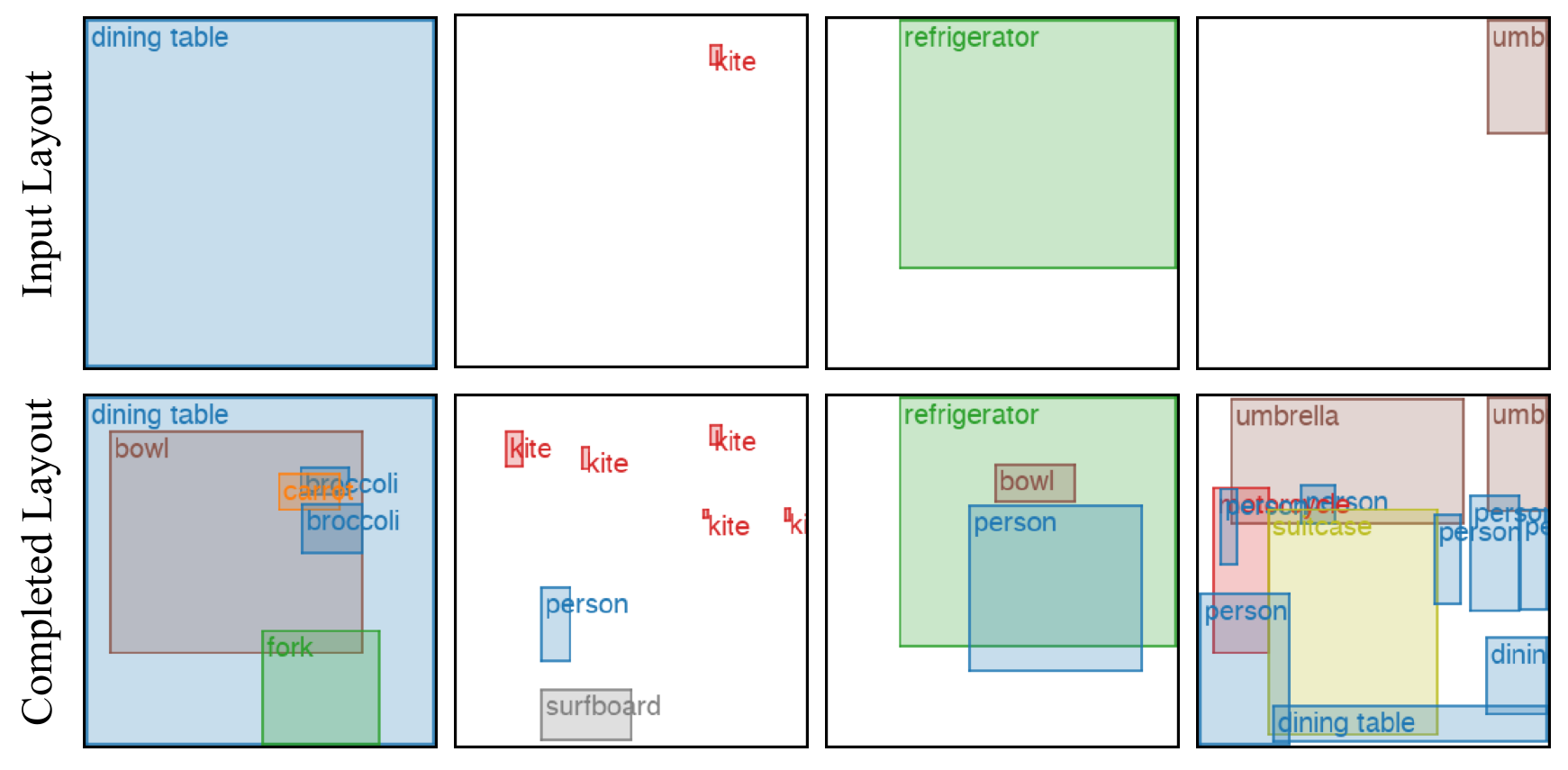}
     \vspace{-0.25in}
     \caption{\textbf{Generated layouts.} Top row shows seed layouts input to the model. Bottom row shows the layout obtained with nucleus sampling. We skip the `stuff' bounding boxes for clarity.}
     \vspace{-0.2in}
    \label{fig:coco_generated}
\end{figure}

\subsection{Layouts for natural scenes}
\label{sec:coco_exp}
COCO bounding boxes dataset is obtained using bounding box annotations in COCO Panoptic 2017 dataset~\citep{lin2014microsoft}. We ignore the images where the \textit{isCrowd} flag is true following the LayoutVAE~\citep{jyothi2019layoutvae} approach. The bounding boxes come from all $80$ thing and $91$ stuff categories. Our final dataset has $118280$ layouts from COCO train split with a median length of $42$ elements and $5000$ layouts from COCO valid split with a median length of $33$. We use the official validation split from COCO as test set in our experiments, and use 5\% of the training data as validation.

\noindent\textbf{Baseline Approaches.} We compare our work with 4 previous methods - LayoutGAN\citep{li2019layoutgan}, LayoutVAE~\citep{jyothi2019layoutvae}, ObjGAN~\citep{li2019object}, and sg2im~\citep{johnson2018image}. Since LayoutVAE and LayoutGAN are not open source, we implemented our own version of these baseline. Note that, like many GAN models, LayoutGAN was notoriously hard to train and our implementation (and hence results) might differ from author's implementation despite our best efforts. We were able to reproduce LayoutVAE's results on COCO dataset as proposed in the original paper and train our own models for different datasets. We also re-purpose ObjGAN and sg2im using guidelines mentioned in LayoutVAE.

Although evaluating generative models is challenging, we attempt to do a fair comparison to the best of our abilities. For our model, we keep architecture hyperparameters same across the datasets. For the baselines, we did a grid search over hyperparameters mentioned in the respective works \& chose the best models according to validation loss. Some ablation studies are provided in the appendix.

\noindent\textbf{Generated Samples.} Fig.~\ref{fig:coco_generated} shows layout completion task using our model on COCO dataset. Although the model is trained with all 171 categories, in the figure we only show `thing' categories for clarity. We also use the generated layouts for a downstream application of scene generation~\citep{zhaobo2019layout2im}.

\begin{figure}[t]
        \centering
        \includegraphics[width=\linewidth]{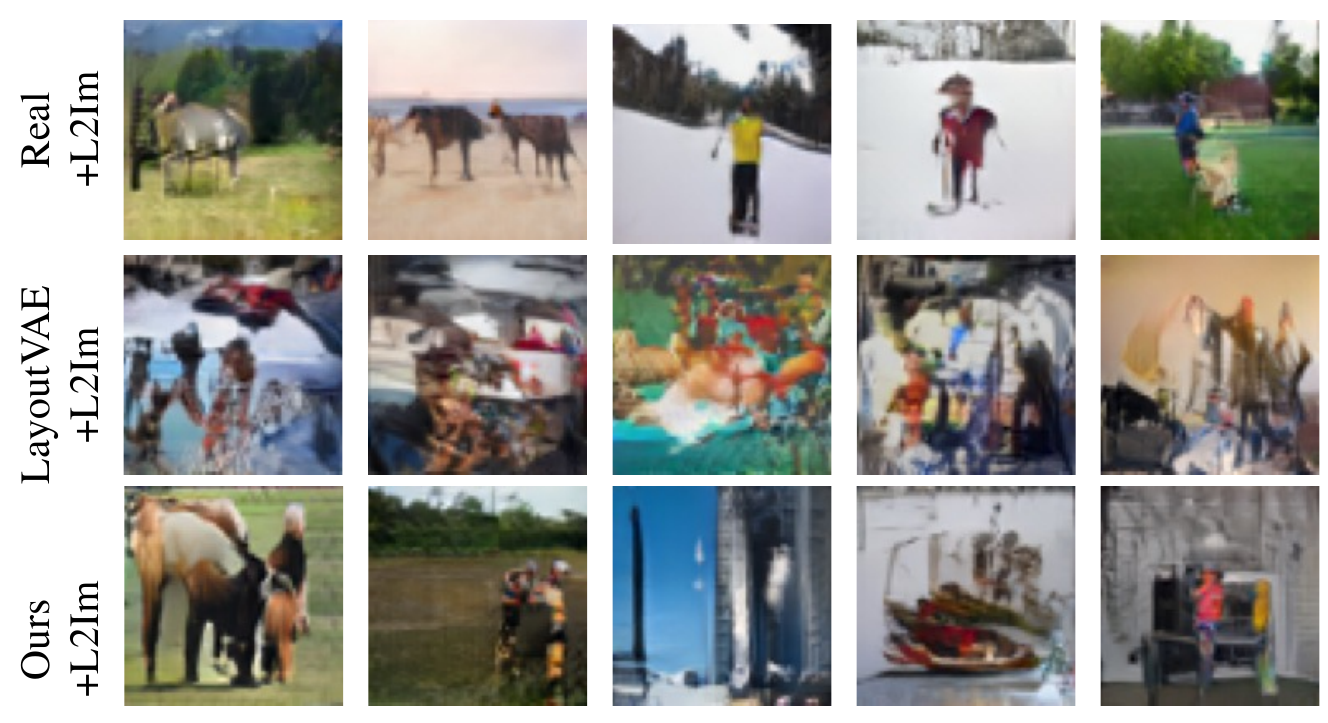}
        \vspace{-0.3in}
        \caption{\textbf{Downstream task.} Image generation with layouts \citep{zhaobo2019layout2im}. FID and IS scores for the generated images provided in Table~\ref{table:fid}.}
        \vspace{-0.2in}
        \label{fig:coco_l2im}
\end{figure}

\begin{figure}[t]
    \centering
     \includegraphics[width=\linewidth]{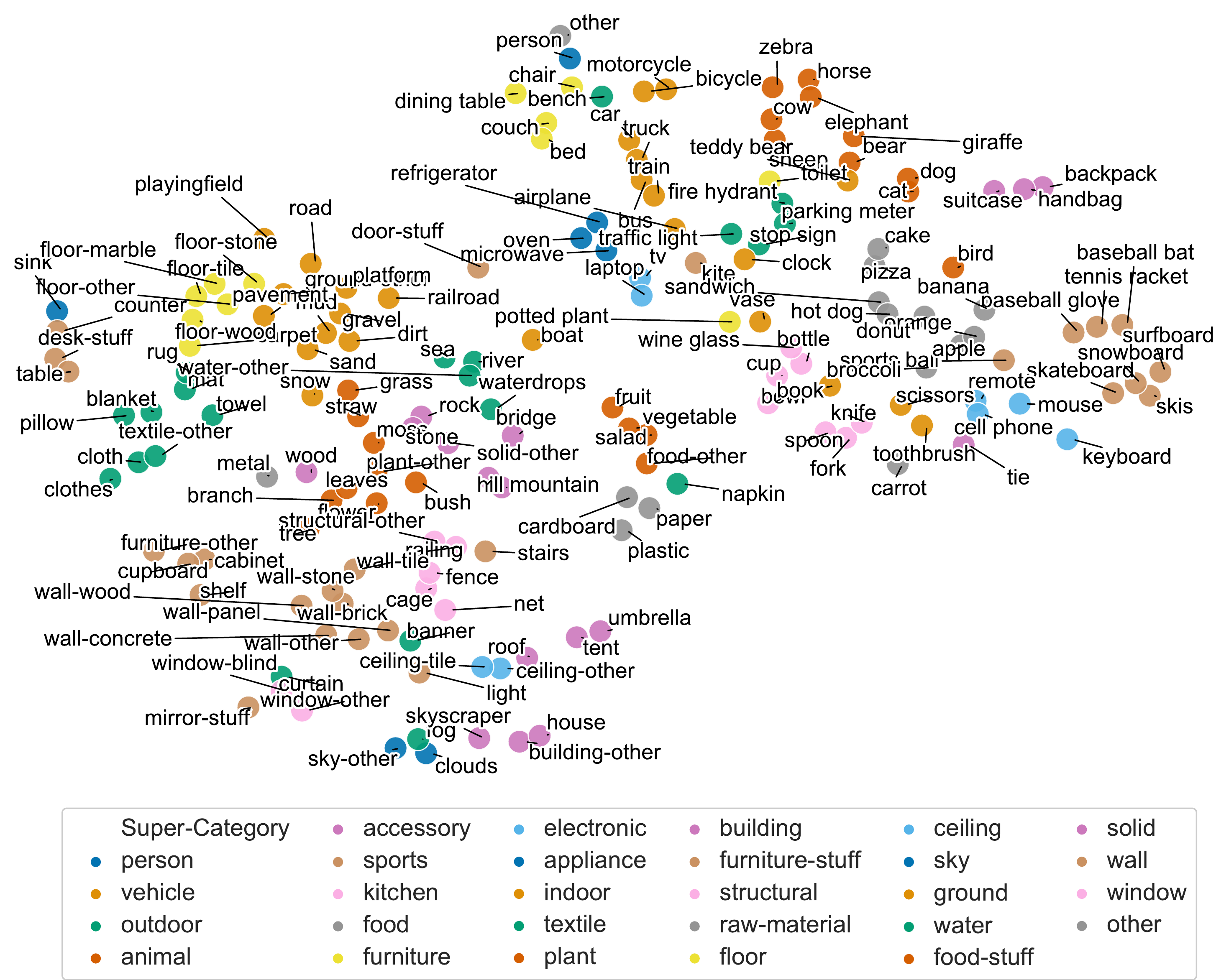}
     \vspace{-0.2in}
      \caption{TSNE plot of learned category embeddings. Words are colored by their super-categories provided in the COCO. Observe that semantically similar categories cluster together. Cats and dogs are closer as compared to sheep, zebra, or cow.}
      \vspace{-0.05in}
    \label{fig:tsne}
\end{figure}

\noindent\textbf{Semantics Emerge via Layout.} We posited earlier that capturing layout should capture contextual relationships between various elements. We provide further evidence of our argument in Fig.~\ref{fig:tsne}. We visualize the 2D-tsne plot of the learned embeddings for categories. We observe that super-categories from COCO are clustered together in the embedding space of the model. Certain categories such as window-blind and curtain (which belong to different super-categories) also appear close to each other. These observations are in line with observations made by ~\cite{gupta2019vico} who use visual co-occurence to learn category embeddings. Table~\ref{table:analogy} shows word2vec~\citep{mikolov2013efficient} style analogies being captured by embeddings learned by our model. Note that the model was trained to generate layouts and we did not specify any additional objective function for analogical reasoning task. 

\begin{figure}[t]
    \centering
        \includegraphics[width=\linewidth]{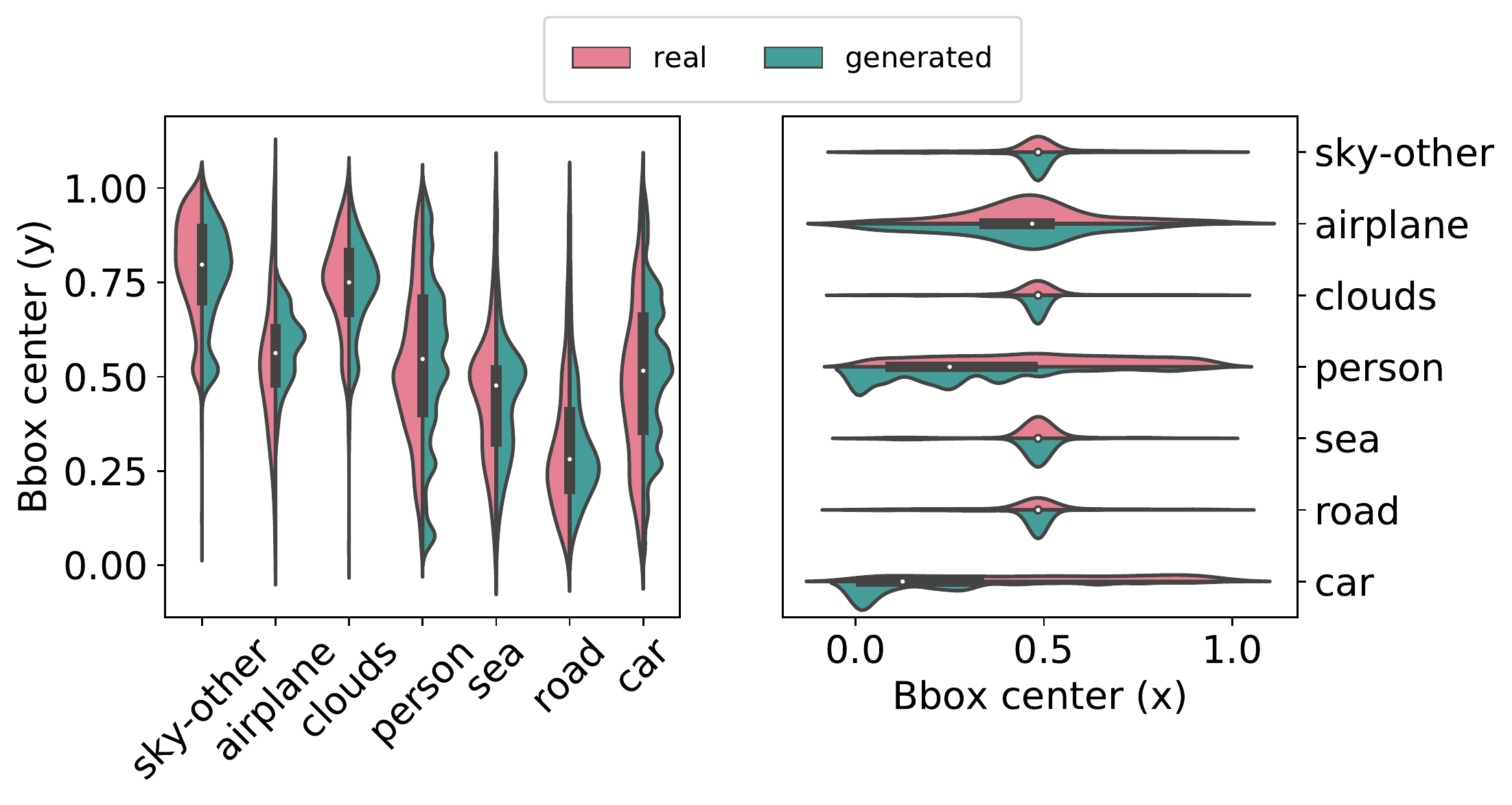}
        \vspace{-0.25in}
        \caption{Distribution of xy-coordinates of bounding boxes centers. Distributions for generated and real layouts is similar. The y-coordinate tends to be more informative (\eg, sky on the top, road and sea at the bottom)}
        \label{fig:violinplot}
    \vspace{-0.05in}
\end{figure}

Finally, we also plot distribution of centers of bounding boxes for various categories in Fig.~\ref{fig:violinplot}. $y-$coordinates of box centers are intuitive since categories such as `sky' or `airplane' are often on top of the image, while `sea' and `road' are at the bottom. This trend is observed in both real and generated layouts. $x-$ coordinates of bounding boxes are more spread out and do not show such a trend. 

\noindent\textbf{Quantitative evaluation.} Following the approach of LayoutVAE, we compute negative log-likelihoods (NLL) of all the layouts in validation data using importance sampling. NLL approach is good for evaluating validation samples, but fails for generated samples. Ideally, we would like to evaluate the performance of a generative model on a downstream task. To this end, we employ Layout2Im~\citep{zhaobo2019layout2im} to generate an image from the layouts generated by each of the method. We compute Inception Score (IS) and Fréchet Inception Distance (FID) to compare quality and diversity of generated images. Our method is competitive with existing approaches in both these metrics, and outperforms existing approaches in terms of NLL.

Note that ObjGAN and LayoutVAE are conditioned on the label set. So we provide labels of objects present in the each validation layout as input. The task for the model is to then predict the number and postition of these objects. Hence, these methods have unfair advantage over our method and ObjGAN indeed performs better than our method and LayoutGAN, which are unconditional. We clearly outperform LayoutGAN on IS and FID metrics.

\begin{table}[t]
\centering
\caption{\textbf{Analogies}. We demonstrate linguistic nuances being captured by our category embeddings by attempting word2vec~\citep{mikolov2013efficient} style analogies.}
\vspace{-0.1in}
\renewcommand{\tabcolsep}{5pt}
\begin{tabular}{@{}c|l @{}}
    \toprule
    Analogy & Nearest neighbors \\
    \midrule
    snowboard:snow::surfboard:?      & waterdrops, sea, sand \\
    car:road::train:?                & railroad, platform, gravel \\
    sky-other:clouds::playingfield:? & net, cage, wall-panel\\
    mouse:keyboard::spoon:?          & knife, fork, oven \\
    fruit:table::flower:?            & potted plant, mirror-stuff \\
    \bottomrule
\end{tabular}
\vspace{-0.15in}
\label{table:analogy}
\end{table}

\subsection{Layouts for Apps and Documents}

\noindent\textbf{Rico Mobile App Wireframes.} Rico mobile app dataset~\citep{Deka:2017:Rico,Liu:2018:LDS:3242587.3242650} consists of layout information of more than $66000$ unique UI screens from over $9300$ android apps. Each layout consists of one or more of the $25$ categories of graphical elements such as text, image, icon, \etc. A complete list elements is provided in the supplementary material. Overall, we get 62951 layouts in Rico with a median length of 36. Since the dataset does not have official splits, we use 5\% of randomly selected layouts for validation and 15\% for testing.

\noindent\textbf{PubLayNet.} PubLayNet~\citep{yepespublaynet} is a large scale document dataset consisting of over $1.1$ million articles collected from PubMed Central. The layouts are annotated with $5$ element categories - text, title, list, label, and figure. We filter out the document layouts with over 128 elements. Our final dataset has $335703$ layouts from official train split with a median length of $33$ elements and $11245$ layouts from dev split with a median length of $36$. We use the dev split as our test set and 5\% of the training data for validation.

\begin{table}[t]
    \centering
    \caption{\textbf{Quantitative Evaluations on COCO.} Negative log-likelihood (NLL) of all the layouts in the validation set (lower the better). We use the importance sampling approach described in ~\cite{jyothi2019layoutvae} to compute. We also generated images from layout using ~\cite{zhaobo2019layout2im} and compute IS and FID. Following~\cite{johnson2018image}, we randomly split test set samples into 5 groups and report standard deviation across the splits. The mean is reported using the combined test set.}
    \vspace{-0.1in}
    \resizebox{0.4\textwidth}{!}{
    \begin{tabular}{@{}lcccc@{}}
        \toprule
        Model & NLL $\downarrow$ & IS $\uparrow$ & FID $\downarrow$ \\
        \midrule
        LayoutGAN~\citep{li2019layoutgan} & - & 3.2 (0.22) & 89.6 (1.6) \\
        LayoutVAE~\citep{jyothi2019layoutvae} & 3.29 & 7.1 (0.41) & 64.1 (3.8) \\
        ObjGAN~\citep{li2019object} & 5.24 & 7.5 (0.44) & 62.3 (4.6) \\
        sg2im~\citep{johnson2018image} & 3.4 & 3.3 (0.15) & 85.8 (1.6) \\
        Ours & \textbf{2.28} & \textbf{7.6 (0.30)} & \textbf{57.0 (3.5)} \\
        \bottomrule
    \end{tabular}}
    \label{table:fid}
    \vspace{-0.15in}
\end{table}

\begin{figure}[t]
        \centering
      \includegraphics[width=\linewidth]{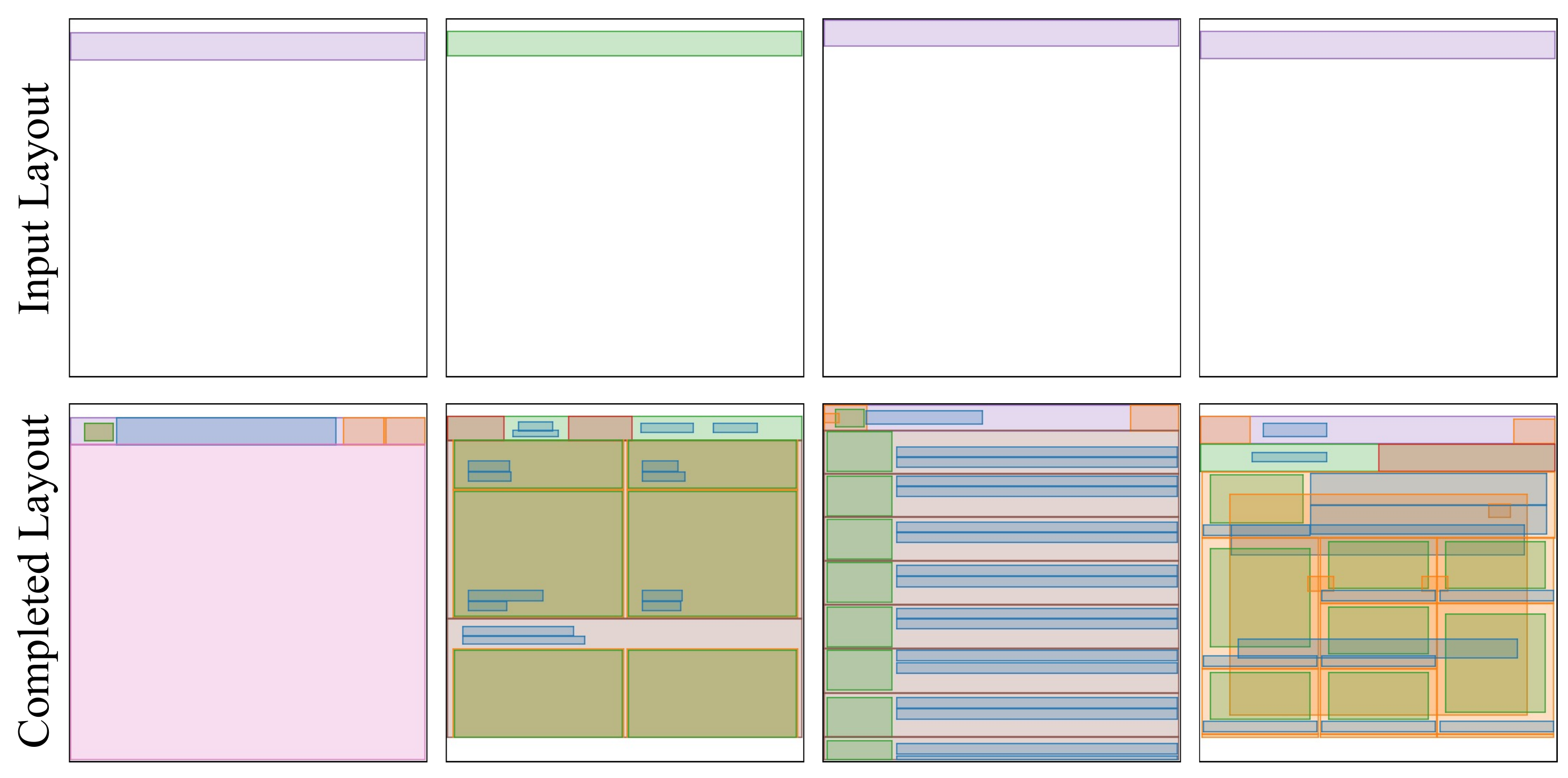}
      \caption{\textbf{RICO layouts.} Generated layouts for the RICO dataset. We skip the categories of bounding boxes for the sake of clarity.}
    \label{fig:rico_generated}
    \vspace{-0.1in}
\end{figure}

\begin{figure}[t]
    \centering
    \includegraphics[width=\linewidth]{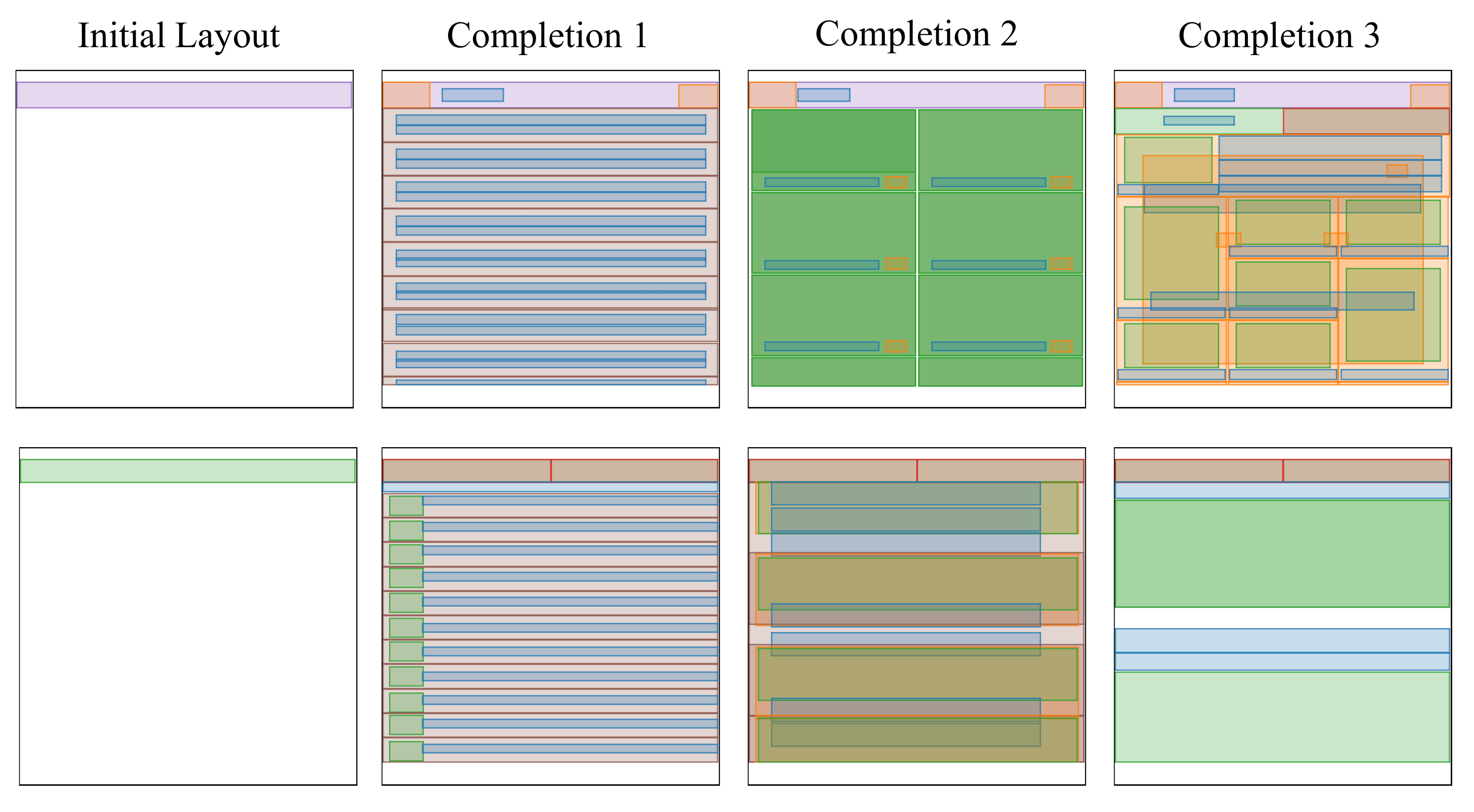}
    \vspace{-0.2in}
    \caption{Multiple completions from same initial element}
    \label{fig:rico_multiple_completions}
    \vspace{-0.2in}
\end{figure}

\begin{figure}[t]
        \centering
        \includegraphics[width=\linewidth]{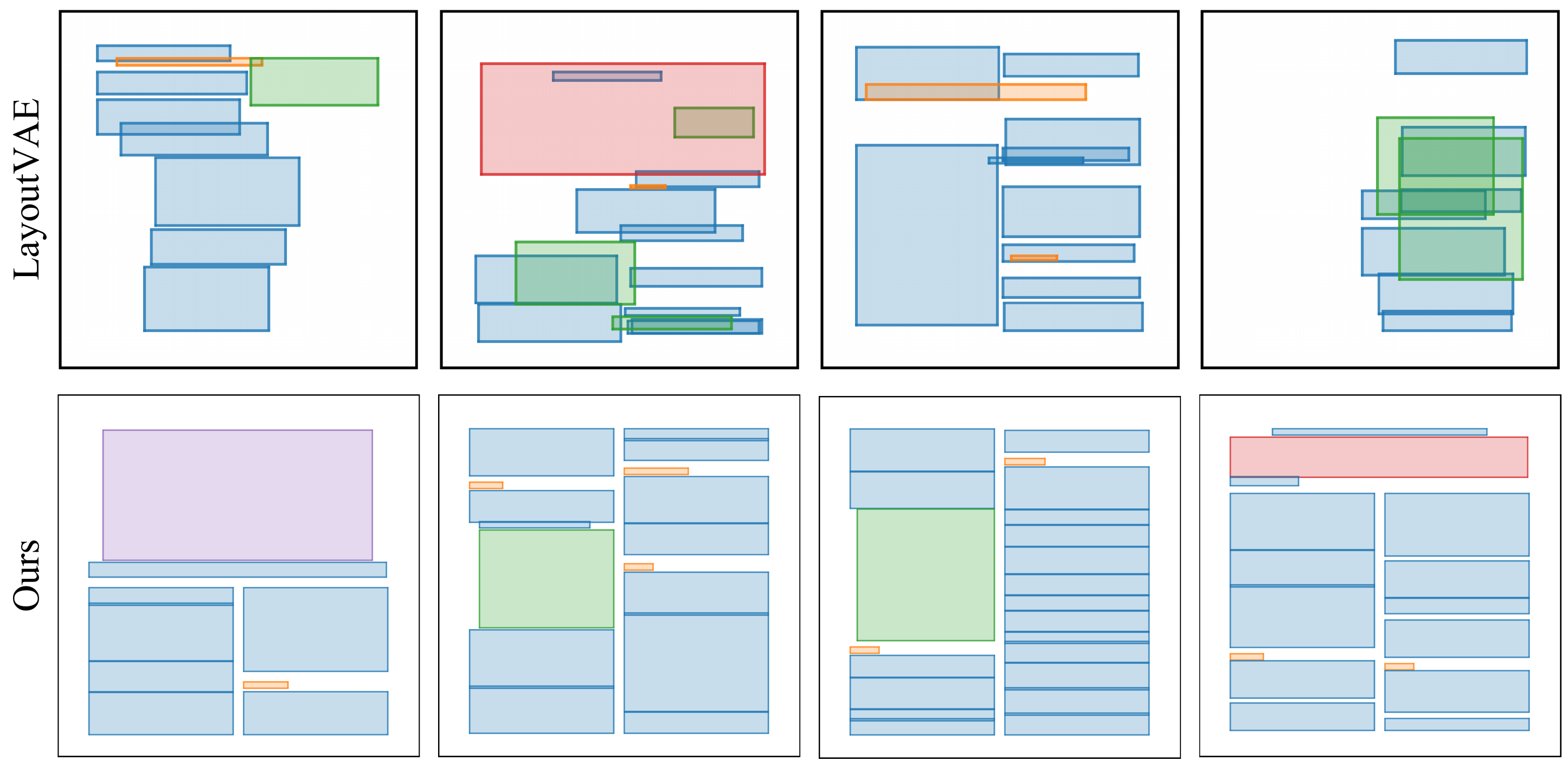}
        \caption{\textbf{Document Layouts.} Generated samples LayoutVAE (top) and our method (bottom). Our method produces aligned bounding boxes for various elements.}
        \label{fig:ours_vs_lv}
\end{figure}

\noindent\textbf{Generated layout samples.} Fig.~\ref{fig:rico_generated} and ~\ref{fig:ours_vs_lv} shows some of the generated samples of our model from RICO mobile app wireframes and PubLayNet documents. Note that both the datasets share similarity in terms of distribution of elements, such as high coverage in terms of space, very little collision of elements, and most importantly alignment of the elements along both x and y-axes. Our method is able to preserve most of these properties as we discuss in the next section.
Fig.~\ref{fig:rico_multiple_completions} shows multiple completions done by our model for the same initial element.

\noindent\textbf{Comparison with baselines.}
We use the same baselines discussed in $\S$\ref{sec:coco_exp}.
Fig.~\ref{fig:ours_vs_lv} shows that our method is able to preserve alignment between bounding boxes better than competing methods. Note that we haven't used any post-processing in order to generate these layouts. Our hypothesis is that (1) discretization of size/position, and (2) decoupling geometric fields in the attention module, are particularly useful in datasets with aligned boxes.

To measure this performance quantitatively, we introduce 2 important statistics. 
\noindent\textbf{Overlap} represents the intersection over union (IoU) of various layout elements. Generally in these datasets, elements do not overap with each other and Overlap is small. \textbf{Coverage} indicates the percentage of canvas covered by the layout elements. Table~\ref{tab:statistics} shows that layouts generated by our method resemble real data statistics better than LayoutGAN and LayoutVAE.

\subsection{Failure Cases}
Our model has a few failure cases, \textit{e.g.}, in Fig.~\ref{fig:shape_generated} in the third object (lamp), the parts are not connected - demonstrating a limitation of our approach arising from training the part auto-encoder and layout generator separately (and not jointly). Similarly, in 2D domains such as COCO, we observe that the model is biased towards generating high frequency categories in the beginning of the generation. This is illustrated in Fig.~\ref{fig:violinplot}, which shows difference in distribution of real \& generated layouts for persons and cars.

\begin{table}[t]
\centering
\caption{Spatial distribution analysis for the samples generated using model trained on RICO and PubLayNet dataset. Closer the Overlap and Coverage values to real data, better is the performance. All values in the table are percentages (std in parenthesis)}
\vspace{-0.08in}
    \resizebox{\linewidth}{!}{
    \renewcommand{\tabcolsep}{8pt}
    \renewcommand{\arraystretch}{1.1}
    \begin{tabular}{@{}lcccccc@{}}
        \toprule
        & \multicolumn{3}{c}{RICO} & \multicolumn{3}{c}{PubLayNet} \\
        \cmidrule(r){2-4}\cmidrule(l){5-7}
        Methods & NLL$\downarrow$ & Coverage & Overlap & NLL$\downarrow$ & Coverage & Overlap. \\
        \midrule
        sg2im~\citep{johnson2018image}          & 7.43 & 25.2 (46) & 16.5 (31) & 7.12 & 30.2 (26) & 3.4 (12) \\
        ObjGAN~\citep{li2019object}             & 4.21 & 39.2 (33) & 36.4 (29) & 4.20 & 38.9 (12) & 8.2 (7) \\
        LayoutVAE~\citep{jyothi2019layoutvae}   & 2.54 & 41.5 (29) & 34.1 (27) & 2.45 & 40.1 (11) & 14.5 (11) \\
        LayoutGAN~\citep{li2019layoutgan}       & -    & 37.3 (31) & 31.4 (32) & -    & 45.3 (19) & 8.3 (10) \\
        Ours                                    & \textbf{1.07} & 33.6 (27) & 23.7 (33) & \textbf{1.10} & 47.0 (12) & 0.13 (1.5) \\
        Real Data                               & -    & 36.6 (27) & 22.4 (32) & -    & 57.1 (10) & 0.1 (0.6) \\
        \bottomrule
    \end{tabular}
    }
\label{tab:statistics}
\vspace{-0.15in}
\end{table}
\section{Conclusion.}
We propose LayoutTransformer, a self-attention framework to generate layouts of graphical elements. Our model uses self-attention model to capture contextual relationship between different layout elements and generate novel layouts, or complete partial layouts. By modeling layout primitives as joint distribution of composable attributes, our model performs competitively with the state-of-the-art approaches for very diverse datasets such as Rico Mobile App Wireframes, COCO bounding boxes, PubLayNet documents, and 3D shapes. We perform a comprehensive qualitative and quantitative evaluation of our model in various domains. We will release our code and models and hope that our model will provide a good starting point for layout modeling applications in various data domains. 

There are a few noteworthy limitations of our approach. First, our model requires a layout or a scene to be decomposed into compositional primitives. In many cases, such primitives might not be even defined. Second, like most data-driven approaches, generated layouts are dominated by high frequency objects or shapes in the dataset. We can control the diversity to some extent using improved sampling techniques, however, generating diverse layouts that not only learn from data, but also from human priors or pre-defined rules is an important direction of research which we will continue to explore.

\medskip
\noindent \textbf{Acknowledgements.} We thank Luis Goncalves, Yuting Zhang, Gowthami Somepalli, and Pulkit Kumar for helpful discussions and reviewing early drafts of the paper. This project was partially funded by the DARPA SAIL-ON (W911NF2020009) program.

\clearpage

{\small
\bibliographystyle{ieee_fullname}
\bibliography{egbib}

\begin{thebibliography}{10}\itemsep=-1pt

\bibitem{ashual2019specifying}
Oron Ashual and Lior Wolf.
\newblock Specifying object attributes and relations in interactive scene
  generation.
\newblock In {\em Proceedings of the IEEE International Conference on Computer
  Vision}, pages 4561--4569, 2019.

\bibitem{ba2016layer}
Jimmy~Lei Ba, Jamie~Ryan Kiros, and Geoffrey~E Hinton.
\newblock Layer normalization.
\newblock {\em arXiv preprint arXiv:1607.06450}, 2016.

\bibitem{biederman1981semantics}
Irving Biederman.
\newblock On the semantics of a glance at a scene.
\newblock {\em Perceptual organization}, 213:253, 1981.

\bibitem{brock2018large}
Andrew Brock, Jeff Donahue, and Karen Simonyan.
\newblock Large scale gan training for high fidelity natural image synthesis.
\newblock {\em arXiv preprint arXiv:1809.11096}, 2018.

\bibitem{DBLP:journals/corr/abs-1710-03474}
Samuele Capobianco and Simone Marinai.
\newblock Docemul: a toolkit to generate structured historical documents.
\newblock {\em CoRR}, abs/1710.03474, 2017.

\bibitem{DBLP:journals/corr/ChangMSPM15}
Angel~X. Chang, Will Monroe, Manolis Savva, Christopher Potts, and
  Christopher~D. Manning.
\newblock Text to 3d scene generation with rich lexical grounding.
\newblock {\em CoRR}, abs/1505.06289, 2015.

\bibitem{chen2020generative}
Mark Chen, Alec Radford, Rewon Child, Jeff Wu, Heewoo Jun, Prafulla Dhariwal,
  David Luan, and Ilya Sutskever.
\newblock Generative pretraining from pixels.
\newblock In {\em Proceedings of the 37th International Conference on Machine
  Learning}, 2020.

\bibitem{chen2019learning}
Zhiqin Chen and Hao Zhang.
\newblock Learning implicit fields for generative shape modeling.
\newblock In {\em Proceedings of the IEEE Conference on Computer Vision and
  Pattern Recognition}, pages 5939--5948, 2019.

\bibitem{Deka:2017:Rico}
Biplab Deka, Zifeng Huang, Chad Franzen, Joshua Hibschman, Daniel Afergan, Yang
  Li, Jeffrey Nichols, and Ranjitha Kumar.
\newblock Rico: A mobile app dataset for building data-driven design
  applications.
\newblock In {\em Proceedings of the 30th Annual Symposium on User Interface
  Software and Technology}, UIST '17, 2017.

\bibitem{dong2017semantic}
Hao Dong, Simiao Yu, Chao Wu, and Yike Guo.
\newblock Semantic image synthesis via adversarial learning.
\newblock In {\em Proceedings of the IEEE International Conference on Computer
  Vision}, pages 5706--5714, 2017.

\bibitem{goodfellow2014generative}
Ian Goodfellow, Jean Pouget-Abadie, Mehdi Mirza, Bing Xu, David Warde-Farley,
  Sherjil Ozair, Aaron Courville, and Yoshua Bengio.
\newblock Generative adversarial nets.
\newblock In {\em Advances in neural information processing systems}, pages
  2672--2680, 2014.

\bibitem{gregor2015draw}
Karol Gregor, Ivo Danihelka, Alex Graves, Danilo~Jimenez Rezende, and Daan
  Wierstra.
\newblock Draw: A recurrent neural network for image generation.
\newblock {\em arXiv preprint arXiv:1502.04623}, 2015.

\bibitem{gupta2020improved}
Kamal Gupta, Susmija Jabbireddy, Ketul Shah, Abhinav Shrivastava, and Matthias
  Zwicker.
\newblock Improved modeling of 3d shapes with multi-view depth maps.
\newblock {\em arXiv preprint arXiv:2009.03298}, 2020.

\bibitem{gupta2020patchvae}
Kamal Gupta, Saurabh Singh, and Abhinav Shrivastava.
\newblock {PatchVAE: Learning Local Latent Codes for Recognition}.
\newblock In {\em Proceedings of the IEEE/CVF Conference on Computer Vision and
  Pattern Recognition}, pages 4746--4755, 2020.

\bibitem{gupta2019vico}
Tanmay Gupta, Alexander Schwing, and Derek Hoiem.
\newblock Vico: Word embeddings from visual co-occurrences.
\newblock In {\em Proceedings of the IEEE International Conference on Computer
  Vision}, pages 7425--7434, 2019.

\bibitem{he2016deep}
Kaiming He, Xiangyu Zhang, Shaoqing Ren, and Jian Sun.
\newblock Deep residual learning for image recognition.
\newblock In {\em Proceedings of the IEEE conference on computer vision and
  pattern recognition}, pages 770--778, 2016.

\bibitem{DBLP:journals/corr/abs-1901-00686}
Tobias Hinz, Stefan Heinrich, and Stefan Wermter.
\newblock Generating multiple objects at spatially distinct locations.
\newblock {\em CoRR}, abs/1901.00686, 2019.

\bibitem{holtzman2019curious}
Ari Holtzman, Jan Buys, Maxwell Forbes, and Yejin Choi.
\newblock The curious case of neural text degeneration.
\newblock {\em arXiv preprint arXiv:1904.09751}, 2019.

\bibitem{hong2018inferring}
Seunghoon Hong, Dingdong Yang, Jongwook Choi, and Honglak Lee.
\newblock Inferring semantic layout for hierarchical text-to-image synthesis.
\newblock In {\em Proceedings of the IEEE Conference on Computer Vision and
  Pattern Recognition}, pages 7986--7994, 2018.

\bibitem{pix2pix2016}
Phillip Isola, Jun-Yan Zhu, Tinghui Zhou, and Alexei~A Efros.
\newblock Image-to-image translation with conditional adversarial networks.
\newblock {\em arxiv}, 2016.

\bibitem{johnson2018image}
Justin Johnson, Agrim Gupta, and Li Fei-Fei.
\newblock Image generation from scene graphs.
\newblock In {\em Proceedings of the IEEE Conference on Computer Vision and
  Pattern Recognition}, pages 1219--1228, 2018.

\bibitem{jyothi2019layoutvae}
Akash~Abdu Jyothi, Thibaut Durand, Jiawei He, Leonid Sigal, and Greg Mori.
\newblock Layoutvae: Stochastic scene layout generation from a label set.
\newblock {\em arXiv preprint arXiv:1907.10719}, 2019.

\bibitem{karras2017progressive}
Tero Karras, Timo Aila, Samuli Laine, and Jaakko Lehtinen.
\newblock Progressive growing of gans for improved quality, stability, and
  variation.
\newblock {\em arXiv preprint arXiv:1710.10196}, 2017.

\bibitem{karras2019style}
Tero Karras, Samuli Laine, and Timo Aila.
\newblock A style-based generator architecture for generative adversarial
  networks.
\newblock In {\em Proceedings of the IEEE Conference on Computer Vision and
  Pattern Recognition}, pages 4401--4410, 2019.

\bibitem{kingma2014adam}
Diederik~P Kingma and Jimmy Ba.
\newblock Adam: A method for stochastic optimization.
\newblock {\em arXiv preprint arXiv:1412.6980}, 2014.

\bibitem{kingma2013autoencoding}
Diederik~P Kingma and Max Welling.
\newblock Auto-encoding variational bayes, 2013.

\bibitem{DBLP:journals/corr/abs-1812-02350}
Donghoon Lee, Sifei Liu, Jinwei Gu, Ming{-}Yu Liu, Ming{-}Hsuan Yang, and Jan
  Kautz.
\newblock Context-aware synthesis and placement of object instances.
\newblock {\em CoRR}, abs/1812.02350, 2018.

\bibitem{li2019seq}
Boren Li, Boyu Zhuang, Mingyang Li, and Jian Gu.
\newblock Seq-sg2sl: Inferring semantic layout from scene graph through
  sequence to sequence learning.
\newblock In {\em Proceedings of the IEEE International Conference on Computer
  Vision}, pages 7435--7443, 2019.

\bibitem{li2019layoutgan}
Jianan Li, Jimei Yang, Aaron Hertzmann, Jianming Zhang, and Tingfa Xu.
\newblock Layoutgan: Generating graphic layouts with wireframe discriminators.
\newblock {\em arXiv preprint arXiv:1901.06767}, 2019.

\bibitem{li2019grains}
Manyi Li, Akshay~Gadi Patil, Kai Xu, Siddhartha Chaudhuri, Owais Khan, Ariel
  Shamir, Changhe Tu, Baoquan Chen, Daniel Cohen-Or, and Hao Zhang.
\newblock Grains: Generative recursive autoencoders for indoor scenes.
\newblock {\em ACM Transactions on Graphics (TOG)}, 38(2):1--16, 2019.

\bibitem{li2019object}
Wenbo Li, Pengchuan Zhang, Lei Zhang, Qiuyuan Huang, Xiaodong He, Siwei Lyu,
  and Jianfeng Gao.
\newblock Object-driven text-to-image synthesis via adversarial training.
\newblock In {\em Proceedings of the IEEE Conference on Computer Vision and
  Pattern Recognition}, pages 12174--12182, 2019.

\bibitem{lin2014microsoft}
Tsung-Yi Lin, Michael Maire, Serge Belongie, James Hays, Pietro Perona, Deva
  Ramanan, Piotr Doll{\'a}r, and C~Lawrence Zitnick.
\newblock Microsoft coco: Common objects in context.
\newblock In {\em European conference on computer vision}, pages 740--755.
  Springer, 2014.

\bibitem{Liu:2018:LDS:3242587.3242650}
Thomas~F. Liu, Mark Craft, Jason Situ, Ersin Yumer, Radomir Mech, and Ranjitha
  Kumar.
\newblock Learning design semantics for mobile apps.
\newblock In {\em The 31st Annual ACM Symposium on User Interface Software and
  Technology}, UIST '18, pages 569--579, New York, NY, USA, 2018. ACM.

\bibitem{gcncnneccv2020}
Dipu Manandhar, Dan Ruta, and John Collomosse.
\newblock Learning structural similarity of user interface layouts using graph
  networks.
\newblock In {\em ECCV}, 2020.

\bibitem{mikolov2013efficient}
Tomas Mikolov, Kai Chen, Greg Corrado, and Jeffrey Dean.
\newblock Efficient estimation of word representations in vector space.
\newblock {\em arXiv preprint arXiv:1301.3781}, 2013.

\bibitem{mo2019structurenet}
Kaichun Mo, Paul Guerrero, Li Yi, Hao Su, Peter Wonka, Niloy Mitra, and
  Leonidas~J Guibas.
\newblock Structurenet: Hierarchical graph networks for 3d shape generation.
\newblock {\em arXiv preprint arXiv:1908.00575}, 2019.

\bibitem{DBLP:journals/corr/abs-1906-02629}
Rafael M{\"{u}}ller, Simon Kornblith, and Geoffrey~E. Hinton.
\newblock When does label smoothing help?
\newblock {\em CoRR}, abs/1906.02629, 2019.

\bibitem{van2016pixel}
Aaron van~den Oord, Nal Kalchbrenner, and Koray Kavukcuoglu.
\newblock Pixel recurrent neural networks.
\newblock {\em arXiv preprint arXiv:1601.06759}, 2016.

\bibitem{park2019deepsdf}
Jeong~Joon Park, Peter Florence, Julian Straub, Richard Newcombe, and Steven
  Lovegrove.
\newblock {DeepSDF: Learning continuous signed distance functions for shape
  representation}.
\newblock In {\em Proceedings of the IEEE Conference on Computer Vision and
  Pattern Recognition}, pages 165--174, 2019.

\bibitem{patil2019read}
Akshay~Gadi Patil, Omri Ben-Eliezer, Or Perel, Hadar Averbuch-Elor, and Cornell
  Tech.
\newblock Read: Recursive autoencoders for document layout generation.
\newblock {\em arXiv preprint arXiv:1909.00302}, 2019.

\bibitem{reed2016generative}
Scott Reed, Zeynep Akata, Xinchen Yan, Lajanugen Logeswaran, Bernt Schiele, and
  Honglak Lee.
\newblock Generative adversarial text to image synthesis.
\newblock {\em arXiv preprint arXiv:1605.05396}, 2016.

\bibitem{ritchie2019fast}
Daniel Ritchie, Kai Wang, and Yu-an Lin.
\newblock Fast and flexible indoor scene synthesis via deep convolutional
  generative models.
\newblock In {\em Proceedings of the IEEE Conference on Computer Vision and
  Pattern Recognition}, pages 6182--6190, 2019.

\bibitem{salimans1701pixelcnn}
Tim Salimans, Andrej Karpathy, Xi Chen, and Diederik~P Kingma.
\newblock Pixelcnn++: Improving the pixelcnn with discretized logistic mixture
  likelihood and other modifications, 2017.
\newblock {\em arXiv preprint arXiv:1701.05517}, 2017.

\bibitem{shrivastava2016contextual}
Abhinav Shrivastava and Abhinav Gupta.
\newblock Contextual priming and feedback for faster r-cnn.
\newblock In {\em European Conference on Computer Vision}, pages 330--348.
  Springer, 2016.

\bibitem{steinbiss1994improvements}
Volker Steinbiss, Bach-Hiep Tran, and Hermann Ney.
\newblock Improvements in beam search.
\newblock In {\em Third International Conference on Spoken Language
  Processing}, 1994.

\bibitem{sung2017complementme}
Minhyuk Sung, Hao Su, Vladimir~G Kim, Siddhartha Chaudhuri, and Leonidas
  Guibas.
\newblock Complementme: weakly-supervised component suggestions for 3d
  modeling.
\newblock {\em ACM Transactions on Graphics (TOG)}, 36(6):1--12, 2017.

\bibitem{szegedy2016rethinking}
Christian Szegedy, Vincent Vanhoucke, Sergey Ioffe, Jon Shlens, and Zbigniew
  Wojna.
\newblock Rethinking the inception architecture for computer vision.
\newblock In {\em Proceedings of the IEEE conference on computer vision and
  pattern recognition}, pages 2818--2826, 2016.

\bibitem{torralba2001statistical}
Antonio Torralba and Pawan Sinha.
\newblock Statistical context priming for object detection.
\newblock In {\em Proceedings Eighth IEEE International Conference on Computer
  Vision. ICCV 2001}, volume~1, pages 763--770. IEEE, 2001.

\bibitem{vaswani2017attention}
Ashish Vaswani, Noam Shazeer, Niki Parmar, Jakob Uszkoreit, Llion Jones,
  Aidan~N Gomez, {\L}ukasz Kaiser, and Illia Polosukhin.
\newblock Attention is all you need.
\newblock In {\em Advances in neural information processing systems}, pages
  5998--6008, 2017.

\bibitem{vinyals2015order}
Oriol Vinyals, Samy Bengio, and Manjunath Kudlur.
\newblock Order matters: Sequence to sequence for sets.
\newblock {\em arXiv preprint arXiv:1511.06391}, 2015.

\bibitem{vondrick2016generating}
Carl Vondrick, Hamed Pirsiavash, and Antonio Torralba.
\newblock Generating videos with scene dynamics.
\newblock In {\em Advances in neural information processing systems}, pages
  613--621, 2016.

\bibitem{wang2019planit}
Kai Wang, Yu-An Lin, Ben Weissmann, Manolis Savva, Angel~X Chang, and Daniel
  Ritchie.
\newblock Planit: Planning and instantiating indoor scenes with relation graph
  and spatial prior networks.
\newblock {\em ACM Transactions on Graphics (TOG)}, 38(4):1--15, 2019.

\bibitem{wang2018deep}
Kai Wang, Manolis Savva, Angel~X Chang, and Daniel Ritchie.
\newblock Deep convolutional priors for indoor scene synthesis.
\newblock {\em ACM Transactions on Graphics (TOG)}, 37(4):70, 2018.

\bibitem{wang2018pix2pixHD}
Ting-Chun Wang, Ming-Yu Liu, Jun-Yan Zhu, Andrew Tao, Jan Kautz, and Bryan
  Catanzaro.
\newblock High-resolution image synthesis and semantic manipulation with
  conditional gans.
\newblock In {\em Proceedings of the IEEE Conference on Computer Vision and
  Pattern Recognition}, 2018.

\bibitem{vda}
Jiajun Wu, Erika Lu, Pushmeet Kohli, William~T Freeman, and Joshua~B Tenenbaum.
\newblock Learning to see physics via visual de-animation.
\newblock In {\em Advances in Neural Information Processing Systems}, 2017.

\bibitem{nsd}
Jiajun Wu, Joshua~B Tenenbaum, and Pushmeet Kohli.
\newblock Neural scene de-rendering.
\newblock In {\em IEEE Conference on Computer Vision and Pattern Recognition
  (CVPR)}, 2017.

\bibitem{wu2016learning}
Jiajun Wu, Chengkai Zhang, Tianfan Xue, Bill Freeman, and Josh Tenenbaum.
\newblock Learning a probabilistic latent space of object shapes via 3d
  generative-adversarial modeling.
\newblock In {\em Advances in neural information processing systems}, pages
  82--90, 2016.

\bibitem{wu2020pq}
Rundi Wu, Yixin Zhuang, Kai Xu, Hao Zhang, and Baoquan Chen.
\newblock Pq-net: A generative part seq2seq network for 3d shapes.
\newblock In {\em Proceedings of the IEEE/CVF Conference on Computer Vision and
  Pattern Recognition}, pages 829--838, 2020.

\bibitem{yang2019pointflow}
Guandao Yang, Xun Huang, Zekun Hao, Ming-Yu Liu, Serge Belongie, and Bharath
  Hariharan.
\newblock {PointFlow: 3D point cloud generation with continuous normalizing
  flows}.
\newblock In {\em Proceedings of the IEEE International Conference on Computer
  Vision}, pages 4541--4550, 2019.

\bibitem{DBLP:journals/corr/YangYAKKG17}
Xiao Yang, Mehmet~Ersin Y{\"{u}}mer, Paul Asente, Mike Kraley, Daniel Kifer,
  and C.~Lee Giles.
\newblock Learning to extract semantic structure from documents using
  multimodal fully convolutional neural network.
\newblock {\em CoRR}, abs/1706.02337, 2017.

\bibitem{yu2019partnet}
Fenggen Yu, Kun Liu, Yan Zhang, Chenyang Zhu, and Kai Xu.
\newblock Partnet: A recursive part decomposition network for fine-grained and
  hierarchical shape segmentation.
\newblock In {\em Proceedings of the IEEE Conference on Computer Vision and
  Pattern Recognition}, pages 9491--9500, 2019.

\bibitem{zhang2018self}
Han Zhang, Ian Goodfellow, Dimitris Metaxas, and Augustus Odena.
\newblock Self-attention generative adversarial networks.
\newblock {\em arXiv preprint arXiv:1805.08318}, 2018.

\bibitem{zhang2017stackgan}
Han Zhang, Tao Xu, Hongsheng Li, Shaoting Zhang, Xiaogang Wang, Xiaolei Huang,
  and Dimitris~N Metaxas.
\newblock Stackgan: Text to photo-realistic image synthesis with stacked
  generative adversarial networks.
\newblock In {\em Proceedings of the IEEE International Conference on Computer
  Vision}, pages 5907--5915, 2017.

\bibitem{zhaobo2019layout2im}
Bo Zhao, Lili Meng, Weidong Yin, and Leonid Sigal.
\newblock Image generation from layout.
\newblock In {\em CVPR}, 2019.

\bibitem{zheng2019content}
Xinru Zheng, Xiaotian Qiao, Ying Cao, and Rynson~WH Lau.
\newblock Content-aware generative modeling of graphic design layouts.
\newblock {\em ACM Transactions on Graphics (TOG)}, 38(4):1--15, 2019.

\bibitem{yepespublaynet}
Xu Zhong, Jianbin Tang, and Antonio~Jimeno Yepes.
\newblock Publaynet: largest dataset ever for document layout analysis.
\newblock In {\em 2019 International Conference on Document Analysis and
  Recognition (ICDAR)}, pages 1015--1022. IEEE, 2019.

\bibitem{zou20173d}
Chuhang Zou, Ersin Yumer, Jimei Yang, Duygu Ceylan, and Derek Hoiem.
\newblock 3d-prnn: Generating shape primitives with recurrent neural networks.
\newblock In {\em Proceedings of the IEEE International Conference on Computer
  Vision}, pages 900--909, 2017.

\end{thebibliography}
}

\clearpage
\appendix

{\centering \textbf{\large{Appendix}}}

\section{Architecture and training details}
\label{sec:network}
In all our $\mathbb{R}^2$ experiments, our base model consists of $d=512$, $L=6$, $n_\text{head}=8$, $\text{precision}=8$ and $d_\text{ff}=2048$. We also use a dropout of $0.1$ at the end of each feedforward layer for regularization. We fix the the maximum number of elements in each of the datasets to $128$ which covers over 99.9\% of the layouts in each of the COCO, Rico and PubLayNet datasets. We also used Adam optimizer~\cite{kingma2014adam} with initial learning rate of $10^{-4}$.
We train our model for 30 epochs for each dataset with early stopping based on maximum log likelihood on validation layouts. Our COCO Bounding Boxes model takes about 6 hours to train on a single NVIDIA GTX1080 GPU. Batching matters a lot to improve the training speed. We want to have evenly divided batches, with minimal padding. We sort the layouts by the number of elements and search over this sorted list to use find tight batches for training.

In all our $\mathbb{R}^3$ experiments, we change $d=128$ and $d_\text{ff}=512$, and learning rate to $10^{-5}$.

\section{Ablation studies}
\label{sec:ablations}
We evaluate the importance of different model components with negative log-likelihood on COCO layouts. The ablation studies show the following:

\smallskip
\noindent\textbf{Small, medium and large elements:} NLL of our model for COCO large, medium, and small boxes is 2.4, 2.5, and 1.8 respectively. We observe that even though discretizing box coordinates introduces approximation errors, it later allows our model to be agnostic to large vs small objects.

\smallskip
\noindent\textbf{Varying $\text{precision}$:} Increasing it allows us to generate finer layouts but at the expense of a model with more parameters. Also, as we increase the $\text{precision}$, NLL increases, suggesting that we might need to train the model with more data to get similar performance (Table~\ref{table:ablation_anchors}).
\begin{table}[!h]
\centering
\caption{Effect of$\text{precision}$ on NLL}
\renewcommand{\tabcolsep}{4pt}
\begin{tabular}{@{}l cccc@{}}
\toprule
$n_\text{anchors}$ & \# params & COCO & Rico & PubLayNet \\
\midrule
    $32 \times 32 $  & 19.2 & 2.28 & 1.07 & 1.10 \\
\midrule
$8 \times 8   $   & 19.1 & 1.69 & 0.98 & 0.88 \\
$16 \times 16 $   & 19.2 & 1.97 & 1.03 & 0.95 \\
$64 \times 64 $   & 19.3 & 2.67 & 1.26 & 1.28 \\
$128 \times 128 $ & 19.6 & 3.12 & 1.44 & 1.46 \\
\bottomrule
\end{tabular}
\label{table:ablation_anchors}
\end{table}

\smallskip
\noindent\textbf{Size of embedding:} Increasing the size of the embedding $d$ improves the NLL, but at the cost of increased number of parameters (Table~\ref{table:ablation_d}).

\begin{table}[!h]
\centering
\caption{Effect of $d$ on NLL}
\renewcommand{\tabcolsep}{4pt}
\begin{tabular}{@{}l cccc@{}}
\toprule
d & \# params & COCO & Rico & PubLayNet \\
\midrule
512 & 19.2 & 2.28 & 1.07 & 1.10 \\
\midrule
32  & 0.8  & 2.51 & 1.56 & 1.26 \\
64  & 1.7  & 2.43 & 1.40 & 1.19 \\
128 & 3.6  & 2.37 & 1.29 & 1.57 \\
256 & 8.1  & 2.32 & 1.20 & 1.56 \\
\bottomrule
\end{tabular}
\label{table:ablation_d}
\end{table}

\smallskip
\noindent\textbf{Model depth:} Increasing the depth of the model $L$, does not significantly improve the results (Table~\ref{table:ablation_layers}). We fix the $L=6$ in all our experiments.

\begin{table}[!h]
    \centering
    \caption{Effect of $L$ on NLL}
    \begin{tabular}{@{}l cccc@{}}
    \toprule
    $L$ & \# params & COCO & Rico & PubLayNet \\
    \midrule
    6 & 19.2 & 2.28 & 1.07 & 1.10 \\
    \midrule
    2 & 6.6  & 2.31 & 1.18 & 1.13 \\
    4 & 12.9 & 2.30 & 1.12 & 1.07 \\
    8 & 25.5 & 2.28 & 1.11 & 1.07 \\
    \bottomrule
    \end{tabular}
    \label{table:ablation_layers}
 \end{table}

\renewcommand{\tabcolsep}{2pt}
\begin{table}[!h]
    \centering
    \caption{Effect of other hyperparameters on NLL}
    \begin{tabular}{@{}lll cccc@{}}
    \toprule
    Order & Split-XY & Loss  & \# params & COCO & Rico & PubLayNet \\
    \midrule
    raster & Yes & NLL   & 19.2 & 2.28 & 1.07 & 1.10 \\
    \midrule
    random &     &       & 19.2 & 2.68 & 1.76 & 1.46 \\
           & No  &       & 21.2 & 3.74 & 2.12 & 1.87 \\
           &     & LS    & 19.2 & 1.96 & 0.88 & 0.88 \\
    \bottomrule
    \end{tabular}
    \label{table:ablation_misc}
\end{table}

\begin{figure*}[!ht]
\centering
\includegraphics[width=\linewidth]{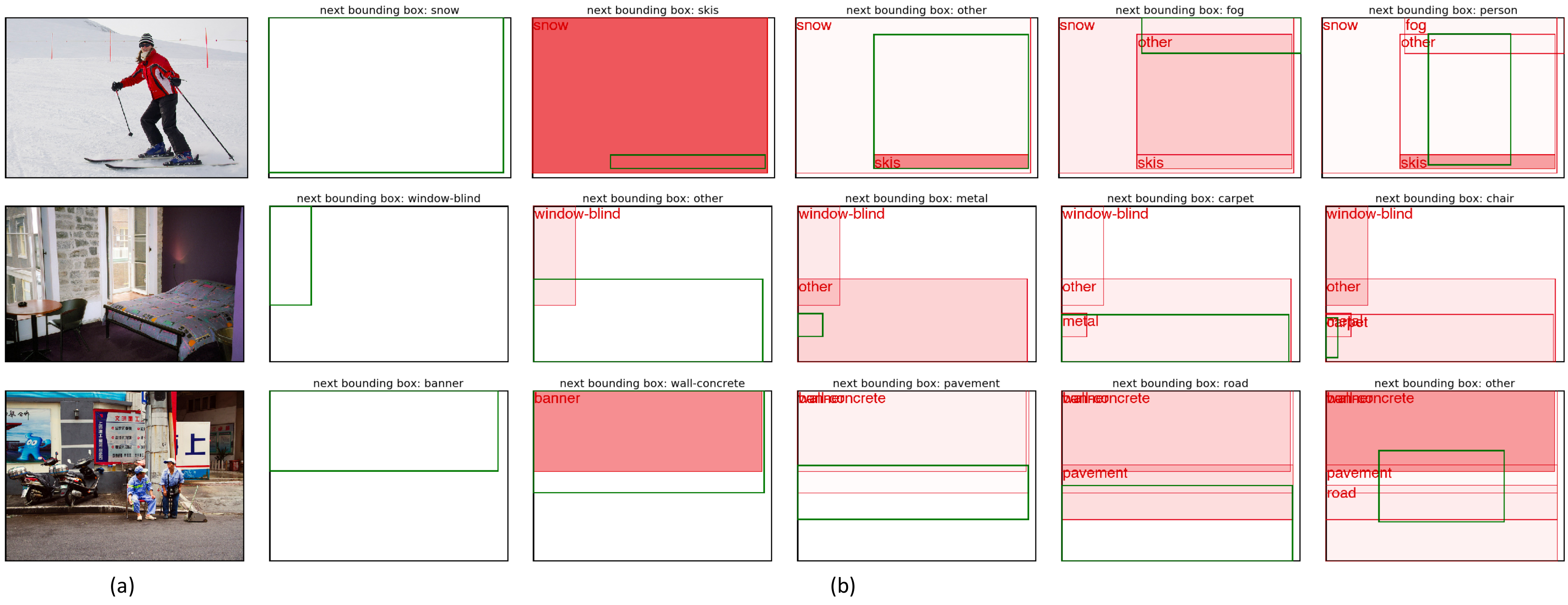}
   \caption{Visualizing attention. (a) Image source for the layout (b) In each row, the model is predicting one element at a time (shown in a green bounding box). While predicting that element, the model pays the most attention to previously predicted bounding boxes (in red). For example, in the first row, ``snow" gets the highest attention score while predicting ``skis". Similarly in the last column, ``skis" get the highest attention while predicting ``person".}
\label{fig:attention}
\end{figure*}

\smallskip
\noindent\textbf{Ordering of the elements:} Adding position encoding, makes the self-attention layer dependent to the ordering of elements. In order to make it depend less on the ordering of input elements, we take randomly permute the sequence. This also enables our model to be able to complete any partial layout. Since output is predicted sequentially, our model is not invariant to the order of output sequence also. In our experiments, we observed that predicting the elements in a simple raster scan order of their position improves the model performance both visually and in terms of negative log-likelihood. This is intuitive as filling the elements in a pre-defined order is an easier problem. We leave the task of optimal ordering of layout elements to generate layouts for future research. (Table~\ref{table:ablation_misc}).

\smallskip
\noindent\textbf{Discretization strategy:} Instead of the factorizing location in x-coordinates and y-coordinates, we tried predicting them at once (refer to the Split-xy column of Table~\ref{table:ablation_misc}). This increases the vocabulary size of the model (since we use $H\times H$ possible locations instead of $H$ alone) and in turn the number of hyper-parameters with decline in model performance. An upside of this approach is that generating new layouts takes less time as we have to make half as many predictions for each element of the layout (Table~\ref{table:ablation_misc}).

\smallskip
\noindent\textbf{Loss:} We tried two different losses, label smoothing~\citep{DBLP:journals/corr/abs-1906-02629} and NLL. Although optimizing using NLL gives better validation performance in terms of NLL (as is expected), we do not find much difference in the qualitative performance when using either loss function. (Table~\ref{table:ablation_misc})

\begin{figure*}[!h]
\centering
    \includegraphics[width=\linewidth]{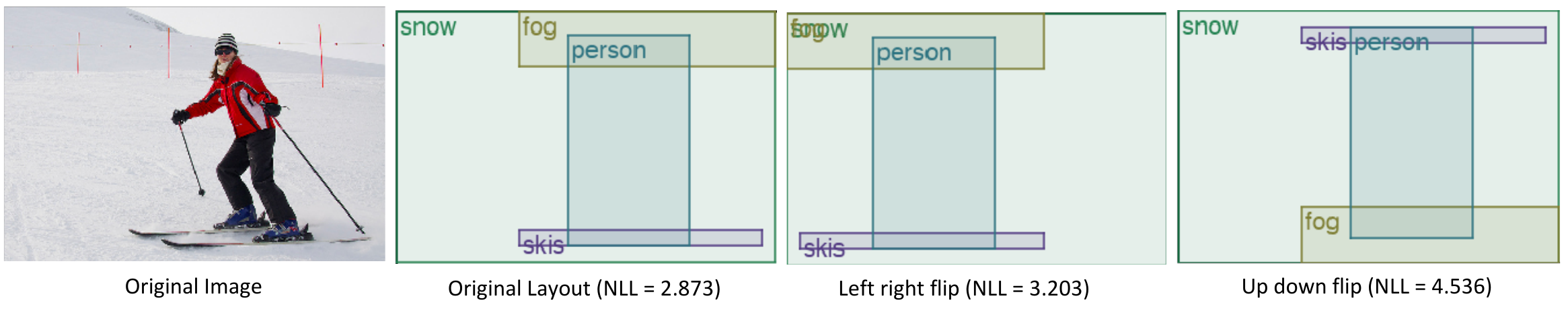}
    \caption{We observe the impact of operations such as left right flip, and up down flip on log likelihood of the layout. We observe that unlikely layouts (such as fog at the bottom of image have higher NLL than the layouts from data.}
\label{fig:nll}
\end{figure*}

\section{Baselines}
\label{sec:evaluation}
In this section we give more details on the various baselines that we implemented or modified from the author's original code. In all the ShapeNet baselines, we used the models provided by the authors or the original numbers provided in the paper. In all the two-dimensional baselines, we implemented the baseline from scratch in the case when code was not available (LayoutVAE) and use the author's own implementation when the code was available (LayoutGAN, ObjGAN, sg2im). Since we train our model in all cases, we ran a grid search over different hyperparameters and report the best numbers in the paper. We found that the GAN based methods were harder to converge and had higher variation in the outcomes as compared to non-adversarial approaches.

\medskip
\noindent\textbf{LayoutVAE.} LayoutVAE~\cite{jyothi2019layoutvae} uses a similar representation for layout and consists of two separate autoregressive VAE models. Starting from a label set, which consists of categories of elements that will be present in a generated layout, their CountVAE generates counts of each of the elements of the label set. After that BoundingBoxVAE, generates the location and size of each occurrence of the bounding box. Note that the model assumes assumption of label set, and hence, while reporting we actually make the task easier for LayoutVAE by providing label-sets of layouts in the validation dataset.

\medskip
\noindent\textbf{ObjGAN.} ObjGAN~\cite{li2019object} provides an object-attention based GAN framework for text to image synthesis. An intermediate step in their image synthesis approach is to generate a bounding box layout given a sentence using a BiLSTM (trained independently). We adopt this step of the ObjGAN framework to our problem setup. Instead of sentences we provide categories of all layout elements as input to the ObjGAN and synthesize all the elements' bounding boxes. This in turn is similar to providing label set as input (similar to the case of LayoutVAE).

\medskip
\noindent\textbf{sg2im.} Image generation from scene graph~\cite{johnson2018image} attempts to generate complex scenes given scene graph of the image by first generating a layout of the scene using graph convolutions and then using the layout to generate complete scene using GANs. The system is trained in an end-to-end fashion. Since sg2im requires a scene graph input, following the approach of ~\cite{johnson2018image}, we create a scene graph from the input and reproduce the input layout using the scene graph.

\medskip
\noindent\textbf{LayoutGAN.} LayoutGAN~\cite{li2019layoutgan} represents each layout with a fixed number of bounding boxes. Starting with bounding box coordinates sampled from a Gaussian distribution, its GAN based framework assigns new coordinates to each bounding box to resemble the layouts from given data. Optionally, it uses non-maximum suppression (NMS) to remove duplicates. The problem setup in LayoutGAN is similar to the proposed approach and they do not condition the generated layout on anything. Note that the authors didn't provide the code for the case of bounding boxes (but only for toy datasets used in the paper). In our implementation, we weren't able to reproduce similar results as the authors reported in the paper for documents.

\section{Visualizing attention}
The self-attention based approach proposed enables us to visualize which existing elements are being attending to while the model is generating a new element. This is demonstrated in Figure~\ref{fig:attention}. We note that While predicting an element, the model pays the most attention to previously predicted bounding boxes (in red). For example, in the first row, ``snow" gets the highest attention score while predicting ``skis". Similarly in the last column, ``skis" get the highest attention while predicting ``person".

\section{Layout Verification}
Since in our method it is straightforward to compute likelihood of a layout, we can use our method to test if a given layout is likely or not. Figure~\ref{fig:nll} shows the NLL given by our model by doing left-right and top-down inversion of layouts in COCO (following ~\cite{li2019layoutgan}). In case of COCO, if we flip a layout left-right, we observe that layout remains likely, however flipping the layout upside decreases the likelihood (or increases the NLL of the layout). This is intuitive since it is unlikely to see fog in the bottom of an image, while skis on top of a person.

\section{More semantics in learned category embeddings}
 Table~\ref{table:bigram} captures the most frequent bigrams and trigrams (categories that co-occur) in real and synthesized layouts. Table~\ref{table:analogy} shows word2vec~\citep{mikolov2013efficient} style analogies being captured by embeddings learned by our model. Note that the model was trained to generate layouts and we did not specify any additional objective function for analogical reasoning task.
 
 \begin{table}[!h]
\centering
\caption{\textbf{Analogies}. We demonstrate linguistic nuances being captured by our category embeddings by attempting to solve word2vec~\citep{mikolov2013efficient} style analogies.}
\renewcommand{\tabcolsep}{5pt}
\begin{tabular}{@{}c|l @{}}
    \toprule
    Analogy & Nearest neighbors \\
    \midrule
    snowboard:snow::surfboard:?      & waterdrops, sea, sand \\
    car:road::train:?                & railroad, platform, gravel \\
    sky-other:clouds::playingfield:? & net, cage, wall-panel\\
    mouse:keyboard::spoon:?          & knife, fork, oven \\
    fruit:table::flower:?            & potted plant, mirror-stuff \\
    \bottomrule
\end{tabular}
\label{table:analogy}
\end{table}

\begin{table*}[!h]
\centering
\caption{\textbf{Bigrams and trigrams}. We consider the most frequent pairs and triplets of (distinct) categories in real \vs generated layouts.}
\renewcommand{\tabcolsep}{5pt}

\begin{tabular}{@{}ll | ll @{}}
    \toprule
    Real & Ours & Real & Ours \\
    \midrule
    other person & other person & person other person & other person clothes \\
    person other & person clothes & other person clothes & person clothes tie \\
    person clothes & clothes tie & person handbag person & tree grass other \\
    clothes person & grass other & person clothes person & grass other person \\
    chair person & other dining table & person chair person & wall-concrete other person \\
    person chair & tree grass & chair person chair & grass other cow \\
    sky-other tree & wall-concrete other & person other clothes & tree other person \\
    car person & person other & person backpack person & person clothes person \\
    person handbag & sky-other tree & person car person & other dining table table \\
    handbag person & clothes person & person skis person & person other person \\
    \bottomrule
\end{tabular}
\label{table:bigram}
\end{table*}

\begin{figure}[!h]
\centering
\includegraphics[width=\linewidth]{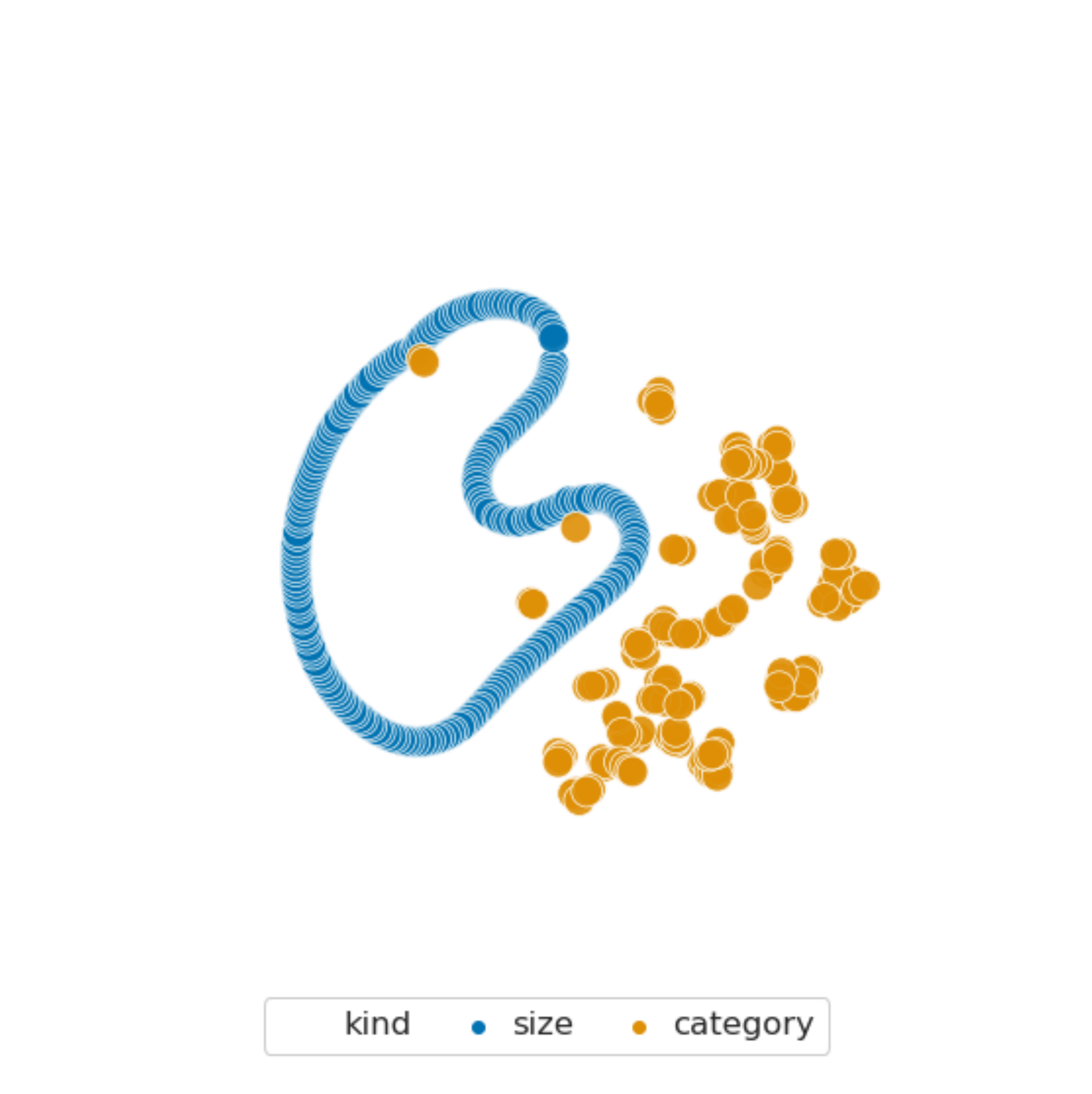}
   \caption{TSNE plot for dimension embedding (256 of them) and category embedding for COCO.}
\label{fig:size_embedding}
\end{figure}

\section{Coordinate Embedding}
Just like in Fig.~\ref{fig:tsne}, we project the embedding learned by our model on COCO in a 2-d space using TSNE. In the absence of explicit constraints on the learned embedding, the model learns to cluster together all the coordinate embedding in a distinct space, in a ring-like manner. Fig.~\ref{fig:size_embedding} shows all the embeddings together in a single TSNE plot.

\section{Nearest neighbors}
\begin{figure*}[!h]
\centering
\includegraphics[width=0.7\linewidth]{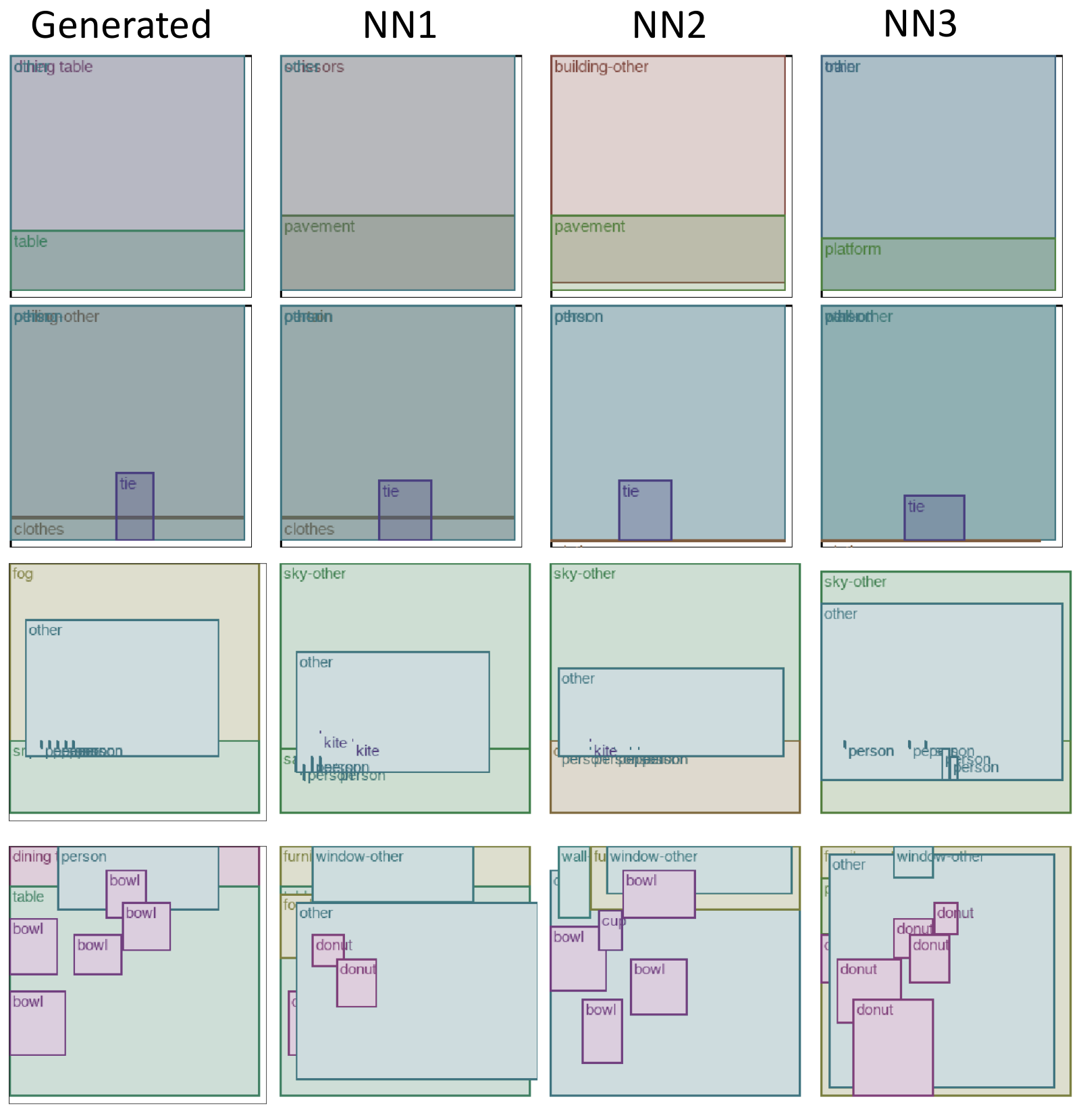}
   \caption{Nearest neighbors from training data. Column 1 shows samples generated by model. Column 2, 3, 4 show the 3 closest neighbors from training dataset. We use chamfer distance on bounding box coordinates to obtain the nearest neighbors from the dataset.}
\vspace{-0.05in}
\label{fig:nn}
\end{figure*}

To see if our model is memorizing the training dataset, we compute nearest neighbors of generated layouts using chamfer distance on top-left and bottom-right bounding box coordinates of layout elements. Figure~\ref{fig:nn}
shows the nearest neighbors of some of the generated layouts from the training dataset. We note that nearest neighbor search for layouts is an active area of research.

\section{More examples for Layout to Image}
\begin{figure*}[!h]
\centering
\includegraphics[width=\linewidth]{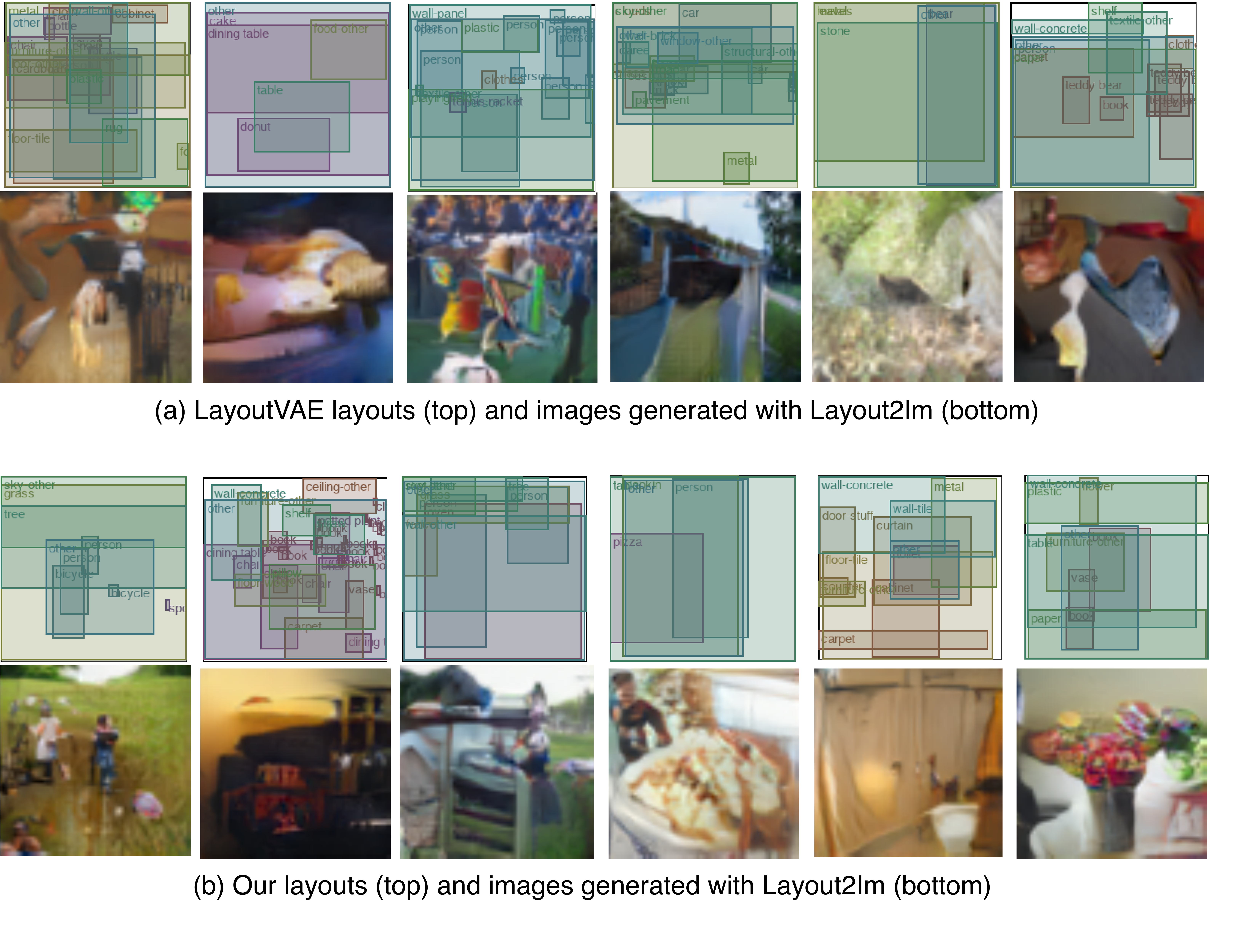}
\caption{Some sample layouts and corresponding images}
\label{fig:l2im_more}
\end{figure*}

Layouts for natural scenes are cluttered and hard to qualitatively evaluate even for a trained user. Here we share some more sample layouts generated from two methods used in the paper. Figure~\ref{fig:l2im_more} shows some extra sample layouts and corresponding images generated using Layout2Im tool. Existing layout to image methods don't work as well as free-form image generation methods but are arguably more beneficial in downstream applications. We hope that improving layout generation will aid the research community to develop better scene generation tools both in terms of diversity and quality.

\section{Dataset Statistics}
\label{sec:datasets}

In this section, we share statistics of different elements and their categories in our dataset. In particular, we share the total number of occurrences of an element in the trai ning dataset (in descending order) and the total number of distinct layouts an element was present in throughout the training data. %
Tables~\ref{table:rico_stats1},~\ref{table:rico_stats2} show the statistics for Rico wireframes, and table~\ref{table:docs_stats} show the statistics for PubLayNet documents.

\begin{table}
\footnotesize
\centering
\caption{Category statistics for Rico}
    \parbox{0.49\textwidth}{
    \centering
    \begin{tabular}{@{}lcc@{}}
    \toprule
    Category & \# occurrences & \# layouts \\
    \midrule
    Text				&	387457	&		50322	\\
    Image				&	179956	&		38958	\\
    Icon				&	160817	&		43380	\\
    Text Button			&	118480	&		33908	\\
    List Item			&	72255	&		9620	\\
    Input				&	18514	&		8532	\\
    Card				&	12873	&		3775	\\
    Web View			&	10782	&		5808	\\
    Radio Button		&	4890	&		1258	\\
    Drawer				&	4138	&		4136	\\
    Checkbox			&	3734	&		1126	\\
    Advertisement		&	3695	&		3365	\\
    \bottomrule
    \end{tabular}
    \label{table:rico_stats1}
    } 
    \hfill
    \parbox{0.49\textwidth}{
    \centering
    \begin{tabular}{@{}lcc@{}}
    \toprule
    Category & \# occurrences & \# layouts \\
    \midrule
    Modal				&	3248	&		3248	\\
    Pager Indicator		&	2041	&		1528	\\
    Slider				&	1619	&		954	\\
    On/Off Switch		&	1260	&		683	\\
    Button Bar		    &	577		&		577	\\
    Toolbar				&	444		&		395	\\
    Number Stepper		&	369		&		147	\\
    Multi-Tab			&	284		&		275	\\
    Date Picker			&	230		&		217	\\
    Map View			&	186		&		94	\\
    Video				&	168		&		144	\\
    Bottom Navigation	&	75		&		27	\\
    \bottomrule
    \end{tabular}
    \label{table:rico_stats2}
    }
\end{table}

\begin{table}[h]
    \centering
    \caption{Category statistics for PubLayNet}
    \begin{tabular}{@{}lcc@{}}
    \toprule
    Category & \# occurrences & \# layouts \\
    \midrule
    text	&	2343356		&		334548	\\
    title	&	627125		&		255731	\\
    figure	&	109292		&		91968	\\
    table	&	102514		&		86460	\\
    list	&	80759		&		53049	\\
    \bottomrule
    \end{tabular}
    \label{table:docs_stats}
\end{table}

\end{document}